\documentclass[10pt,twocolumn,letterpaper]{article}

\usepackage{cvpr}              

\usepackage{graphicx}
\usepackage{amsmath}
\usepackage{amssymb}
\usepackage{booktabs}

\usepackage{algorithm}
\usepackage{algorithmic}

\usepackage{amsmath,amsfonts,bm}

\def\eqref#1{equation~\ref{#1}}

\def\1{\bm{1}}

\DeclareMathAlphabet{\mathsfit}{\encodingdefault}{\sfdefault}{m}{sl}
\SetMathAlphabet{\mathsfit}{bold}{\encodingdefault}{\sfdefault}{bx}{n}
\newcommand{\tens}[1]{\bm{\mathsfit{#1}}}

\def\tK{{\tens{K}}}

\DeclareMathOperator*{\argmax}{arg\,max}

\newtheorem{Definition}{Definition}
\newtheorem{Remark}{Remark}

\newtheorem{Proposition}{Proposition}

\usepackage{bbm}

\usepackage{booktabs}
\usepackage{amsmath}
\usepackage{multirow}
\usepackage{amssymb}
\usepackage{pifont}
\usepackage{xcolor}
\definecolor{mypink}{rgb}{0.858, 0.188, 0.478}

\usepackage[pagebackref,breaklinks,colorlinks]{hyperref}

\usepackage[capitalize]{cleveref}
\crefname{section}{Sec.}{Secs.}
\Crefname{section}{Section}{Sections}
\Crefname{table}{Table}{Tables}
\crefname{table}{Tab.}{Tabs.}

\def\cvprPaperID{3079} 
\def\confName{CVPR}
\def\confYear{2022}

\begin{document}

\title{Category Contrast for Unsupervised Domain Adaptation in Visual Tasks}

\author{Jiaxing Huang \textsuperscript{\rm 1}, Dayan Guan \textsuperscript{\rm 1}, Aoran Xiao \textsuperscript{\rm 1}, Shijian Lu\thanks{Corresponding author.} \  \textsuperscript{\rm 1}, Ling Shao \textsuperscript{\rm 2} \\ 
\textsuperscript{\rm 1} Nanyang Technological University, \textsuperscript{\rm 2} Inception Institute of Artificial Intelligence.\\
{\tt\small \{Jiaxing.Huang, Dayan.Guan, Aoran.Xiao, Shijian.Lu\}@ntu.edu.sg, ling.shao@ieee.org}
}

\maketitle

\begin{abstract}
Instance contrast for unsupervised representation learning has achieved great success in recent years.
In this work, we explore the idea of instance contrastive learning in unsupervised domain adaptation (UDA) and propose a novel \textit{Category Contrast} technique (CaCo) that introduces semantic priors on top of instance discrimination for visual UDA tasks.
By considering instance contrastive learning as a dictionary look-up operation, we construct a semantics-aware dictionary with samples from both source and target domains where each target sample is assigned a (pseudo) category label based on the category priors of source samples.
This allows category contrastive learning (between target queries and the category-level dictionary) for category-discriminative yet domain-invariant feature representations: samples of the same category (from either source or target domain) are pulled closer while those of different categories are pushed apart simultaneously. Extensive UDA experiments in multiple visual tasks 
($e.g.$, segmentation, classification and detection) 
show that CaCo achieves superior performance as compared with state-of-the-art methods. The experiments also demonstrate that CaCo is complementary to existing UDA methods and generalizable to other learning setups such as unsupervised model adaptation, open-/partial-set adaptation etc.
\end{abstract}

\section{Introduction}
Though deep neural networks (DNNs)~\cite{simonyan2014very,he2016deep} have revolutionized various computer vision tasks~\cite{chen2017deeplab,ren2015faster, simonyan2014very,he2016deep}, they often generalize poorly to new domains due to \textit{domain gaps}. Unsupervised domain adaptation (UDA) aims to mitigate the domain gaps by exploiting unlabelled target-domain data. To this end, researchers have designed different unsupervised losses on target data for learning a well-performed model in target domain ~\cite{kang2019contrastive,tsai2018learning, tzeng2017adversarial, luo2019taking,vu2019advent,chen2018domain,xiao2022unsupervised}. The existing unsupervised losses can be broadly classified into three categories: 1) \textit{adversarial loss} that enforces source-like target representations in the feature, output or latent space~\cite{ long2016unsupervised,tzeng2017adversarial, luo2019taking, tsai2018learning,saito2018maximum,vu2019advent,tsai2019domain}; 2) \textit{image translation loss} that translates source images to have target-like styles and appearance \cite{ chen2019crdoco,li2019bidirectional,huang2021rda,yang2020fda,zhang2021spectral}; and 3) \textit{self-training loss} that re-trains networks iteratively with confidently pseudo-labelled target samples \cite{zou2018unsupervised,zou2019confidence, li2019bidirectional,guan2021domain}.

\begin{figure*}
\centering
\includegraphics[width=0.99\linewidth]{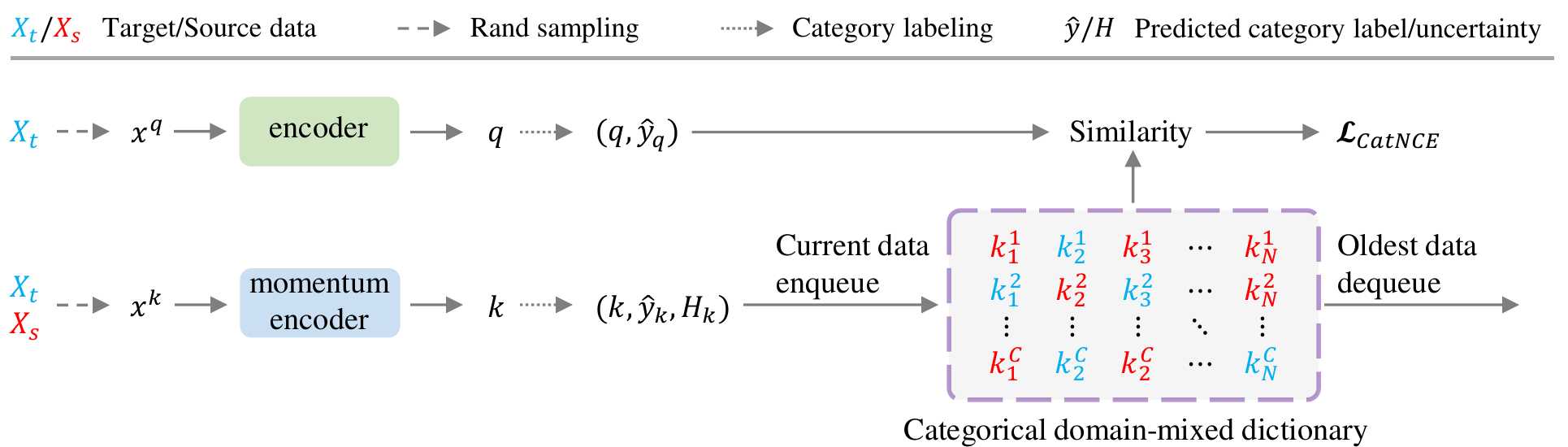}
\caption{
The proposed Category Contrast trains an unsupervised domain adaptive encoder by matching a query $q$ (from an unlabelled target sample $x^q\in X_t$) to a dictionary of keys via a category contrastive loss $\mathcal{L}_\mathrm{CatNCE}$. The dictionary keys are domain-mixed from both source domain $X_s$ (in red with labels) and target domain $X_t$ (in blue with pseudo labels), which allows to learn invariant representations within and across the two domains. They are also category-ware and category-balanced allowing to learn category-discriminative yet category-unbiased representations. Note the \textit{category-balanced} means that each query $q$ is compared with all the dictionary keys (in loss computation) that are evenly distributed over all data categories which mitigates data imbalance issue.
}
\label{fig:framework}
\end{figure*}

Unsupervised representation learning~\cite{noroozi2016unsupervised,pathak2016context,zhang2016colorful,zhang2017split,wu2018unsupervised,ye2019unsupervised,he2020momentum, tian2019contrastive,chen2020simple} addresses a related problem, $i.e.$, unsupervised network pre-training which aims to learn discriminative embeddings from unlabelled data. In recent years, instance contrastive learning~\cite{wu2018unsupervised,ye2019unsupervised,he2020momentum, oord2018representation, tian2019contrastive,chen2020simple} has led to major advances in unsupervised representation learning.
Despite different motivations, instance contrast methods can be thought of as a dictionary look-up task~\cite{he2020momentum} that trains a visual encoder by matching an encoded query $q$ with
a dictionary of encoded keys $k$:
the encoded query should be similar to the encoded positive keys and dissimilar to encoded negative keys.
With no labels available for unlabelled data, the positive keys are often the augmentation of the query sample, and all the rest are negative keys.

In this work, we explore the idea of instance contrast
in UDA. Considering contrastive learning as a dictionary look-up task, we hypothesize that a UDA dictionary should be category-aware and domain-mixed with keys from both source and target domains. Intuitively, a category-aware dictionary with category-balanced keys will encourage to learn category-discriminative yet category-unbiased representations, while the keys from both source and target domains will allow to learn invariant representations within and across the two domains, both being aligned with the objective of UDA.

With above motivation, this paper presents \textit{Category Contrast} (CaCo) as a way of building category-aware and domain-mixed dictionaries with corresponding contrastive losses for UDA. As illustrated in Fig.~\ref{fig:framework}, the dictionary consists of keys that are evenly sampled in both categories and domains, where each target key comes with a predicted pseudo category. Take the illustrative dictionary $\tK = \{k^{c}_{m}\}_{1 \leq c \leq C, 1 \leq m \leq M}$ as an example. Each category $c$ will have $M$ keys while each domain has $(C\times M)/2$ keys. The network learning will thus strive to minimize a \textit{category contrastive loss} $\mathcal{L}_{\text{CatNCE}}$ between target queries and dictionary keys: samples of the same category are pulled close while those of different categories are pushed away.
This naturally leads to category-discriminative yet domain-invariant representations, which is perfectly aligned with the objective of UDA.

With the category-aware and domain-mixed dictionary together with the category contrastive loss, the proposed Category Contrast tackles the UDA challenges with three desirable features: 1) It concurrently minimizes the intra-category variation and maximizes the inter-category distance with the \textit{category-aware} dictionary design; 2) It achieves inter-domain and intra-domain alignment simultaneously thanks to the \textit{domain-mixed} dictionary design by including both source and target samples; 3) It greatly mitigates the data balance issue due to the \textit{category-balanced} dictionary design which allows to compute contrast losses evenly across all categories during learning. 

The contributions of this work can be summarized in three aspects. 
\textit{First}, we investigate instance contrast for unsupervised domain adaptation, aiming to learn discriminative representation for unlabelled target-domain data. 
\textit{Second}, we propose Category Contrast that builds a category-aware and domain-mixed dictionary with a category contrastive loss. It encourages to learn category-discriminative yet domain-invariant representation which is naturally aligned with the objective of UDA. \textit{Third}, extensive experiments show that the proposed CaCo achieves superior UDA performance consistently as compared with state-of-the-art methods.
Additionally, CaCo is complementary with existing UDA methods and generalizable to other learning setups that involves unlabeled data.

\section{Related Works}

Our work is closely related to two main branches of research, namely, unsupervised learning in unsupervised domain adaptation and instance contrast in unsupervised representation learning.

\textbf{Unsupervised domain adaptation} aims to leverage unlabelled target data to improve network performance in target domain. To learn from unlabelled target data, most existing works propose different unsupervised losses which can be broadly classified into three categories. The first category is \textit{adversarial loss} that enforces source-like target representation in the feature~\cite{long2016unsupervised,tzeng2017adversarial,chen2018domain,guan2021uncertainty,saito2019strong,zhang2021detr}, output~\cite{saito2018adversarial,luo2019taking,tsai2018learning,saito2018maximum,huang2022multi} or latent space~\cite{vu2019advent,tsai2019domain,huang2020contextual}. The second category is \textit{image translation loss} that generates source data with target-like styles and appearance via GANs~\cite{chen2019crdoco, li2019bidirectional,cui2021genco} and spectrum matching~\cite{yang2020fda,huang2021fsdr}. The third category is \textit{self-training loss} that re-trains the network iteratively with pseudo-labelled target samples \cite{zou2018unsupervised, zou2019confidence,li2019bidirectional,guan2021scale,huang2021model,yang2020fda,huang2021cross,wang2021domain}.

We tackle UDA from a new perspective of instance contrastive learning, and propose a novel Category Contrast (CaCo) that introduces a generic category contrastive loss that can work for various UDA tasks.
To the best of our knowledge, this is the first work that explores instance contrastive learning for UDA.

\textbf{Instance Contrastive Learning}~\cite{wu2018unsupervised,ye2019unsupervised, he2020momentum, oord2018representation, tian2019contrastive,chen2020simple} aims to learn an embedding space where positive samples are pulled close to an anchor and negative samples are pushed away. Despite different motivations, instance contrastive learning can be viewed as a dictionary look-up task~\cite{he2020momentum} that trains a visual encoder by matching an encoded query $q$ with a dictionary of encoded keys $k$: $q$ should be similar to positive $k$ and dissimilar to negative $k$. Three typical dictionary creation strategies have been proposed. The first builds a \textit{memory bank}~\cite{wu2018unsupervised} that stores the keys of all samples in the previous epoch. The second creates an \textit{end-to-end} dictionary~\cite{ye2019unsupervised, tian2019contrastive, chen2020simple} that generates keys from samples of the current mini-batch. The third employs a \textit{momentum encoder}~\cite{he2020momentum} that encodes samples on-the-fly by a momentum-updated encoder. Instance contrastive learning with various dictionaries helps to learn better unsupervised representations clearly.

On the other hand, existing instance contrastive learning methods~\cite{wu2018unsupervised,ye2019unsupervised,he2020momentum,oord2018representation,tian2019contrastive,chen2020simple} were designed for unsupervised representation, which has two main limitations in UDA: 1). With little category priors, existing instance contrast techniques learn rich low-level features without capturing much high-level semantic information. This is sub-optimal to many visual recognition tasks ($e.g.$, segmentation, detection and classification) that require discriminative semantic features. Recent studies~\cite{tschannen2019mutual,saunshi2019theoretical} verify this issue; 2). Most existing instance contrastive learning methods~\cite{wu2018unsupervised, ye2019unsupervised,he2020momentum,oord2018representation,tian2019contrastive,chen2020simple} employ a super-large/category-agnostic dictionary that could introduce category collision~\cite{saunshi2019theoretical}, where negative pairs share the same semantic category but are undesirably pushed away in the feature space. This impairs most learning setups that require  semantic-level discrimination including various visual UDA tasks. The proposed CaCo introduces a categorical domain-mixed dictionary which introduces category priors and addresses the two problems effectively.

\textbf{Other recent related contrastive learning works.} \cite{li2021semantic} explores contrastive learning with semantic distributions and proposes semantic distribution-aware contrastive adaptation that contrasts each sample with estimated category centroids. \cite{wang2021exploring,alonso2021semi} explore pixel-level contrast with a memory bank for supervised and semi-supervised semantic segmentation. 

\section{Method}

\subsection{Task Definition}

This work focuses on the task of unsupervised domain adaptation. Given labeled source-domain data \{$X_{s}$, $Y_{s}$\} and unlabeled target-domain data $X_{t}$, the goal is to learn a model $G$ that performs well over $X_{t}$. The \textit{baseline} model is trained with the labeled source data only:
\begin{equation}
\begin{split}
\mathcal{L}_{sup} = l(G(X_{s}), Y_{s}),    
\end{split}
\label{eq_baseline}
\end{equation}
where $l(\cdot)$ denotes an accuracy-related loss, $e.g.$, the standard cross-entropy loss.

\subsection{Preliminaries of Instance Contrastive Learning}
\label{sec:preliminaries}

The idea of instance contrastive learning \cite{hadsell2006dimensionality} can be considered as training an encoder (feature extractor) for a \emph{dictionary look-up} task.
Given a query $q$ and a dictionary that consists of a number of keys $\{k_{0}, k_{1}, ..., k_{N}\}$, instance discriminative representations are learnt with an instance contrastive loss \cite{hadsell2006dimensionality} ($e.g.$, InfoNCE \cite{oord2018representation}), minimization of which will pull $q$ close to its positive key and push it away from all other keys (considered negative for $q$):
\begin{equation}
\mathcal{L}_{\text{InfoNCE}} = \sum_{x_{q} \in X} -\log \frac{\sum_{i=0}^{N} \mathbbm{1} (k_{i} \in q) \exp(q{\cdot}k_{i} / \tau)}{\sum_{i=0}^{N}\exp(q{\cdot}k_i  / \tau)}
\label{eqn:infoNCE}
\end{equation}
where $\mathbbm{1} (k_{i} \in q) = 1$ if $k_{i}$ is the positive key of $q$ and $ \mathbbm{1} (k_{i} \in q) = 0$ otherwise. Parameter $\tau$ is a temperature parameter \cite{wu2018unsupervised}. In general, the query representation is $q=f_\textrm{q}(x^q)$ where $f_\textrm{q}$ is an encoder network and $x^q$ is a query sample (likewise in $k=f_\textrm{k}(x^k)$).

\subsection{Category Contrast for Unsupervised Domain Adaptation}

We tackle UDA from a perspective of instance contrastive learning. Specifically, we design Category Contrast that builds a category-aware and domain-mixed dictionary to learn category-discriminative yet domain-invariant representations under the guidance of a category contrastive loss.

\textbf{Overview.} For \textit{supervised training} over a labelled source domain, we feed source samples $\{X_{s}, Y_{s}\}$ to a model $G$ and optimize $G$ with Eq.~\ref{eq_baseline}. In this work, $G$ consists of an encoder $f_{q}$ and a classifier $h$ that classifies the encoded embeddings into pre-defined categories, $i.e.$, $G(\cdot)=h(f_{q}(\cdot))$. For \textit{unsupervised training} over an unlabelled target domain, the training involves a query \textit{encoder} $f_{q}$ and a key \textit{momentum encoder} $f_{k}$ (the momentum update of $f_{q}$, $i.e.$, $\theta_{f_{k}} = b\theta_{f_{k}} + (1- b)\theta_{f_{q}}$, and $b$ is a momentum coefficient) as illustrated in Fig.~\ref{fig:framework}. During the training, we evenly sample the key $x_{k}$ from both source and target domains ($i.e.$, $X_{s}$ and $X_{t}$) and feed them to the key encoder $f_{k}$ to build a category-aware dictionary $\tK$. We sample query $x_{q}$ from the target domain (i.e. $X_{t}$) only and feed them to the query encoder $f_{q}$ for category contrastive learning with the category-aware dictionary $\tK$.

\textbf{Categorical domain-mixed dictionary.} One key component in the proposed CaCo is a category-aware and domain-mixed dictionary with keys from both source and target domains. The dictionary allows to perform category contrastive learning: the embeddings of the \textit{same category} are pulled close together while those of \textit{different categories} are pushed apart. The category awareness encourages the network to learn category-discriminative embeddings. This feature is critical to various visual tasks ($e.g.$, segmentation, classification and detection) that require to learn discriminative features and classify them to pre-defined categories. In addition, the dictionary is domain-mixed which encourages to learn invariant representations within and across domains as category contrast is computed between target queries and keys from both source and target domains.

As stated in the Overview, given an encoded key $k = f_{k}(x_k)$ ($x_{k} \in X_{s} \cup X_{t}$), the classifier $h$ predicts a category label $\hat{y}_{k}$ and converts $k$ into a categorical key $k^{c}$ which is further queued into the categorical dictionary $\tK$. These processes are carried out in parallel for a mini-batch of inputs, and the formal definition of the categorical dictionary $\tK$ is presented in Definition.~\ref{CaCo-definition}.

\begin{Definition}
\label{CaCo-definition} A Categorical Dictionary $\tK$ with $C$-category is defined by:
\begin{equation}
\label{CaCo-Dict}
\tK = \{k^{1}, k^{2}, ..., k^{C}\},
\end{equation}
where the categorical key $k^{c} \in \tK$ is defined as the key $k$ that belongs to the $c$-th semantic category ($c = \argmax_{i}\hat{y}_{k}^{(i)}$) and the predicted category label $\hat{y}_{k}$ of $k = f_{k}(x_{k})$ is derived by:
\begin{equation}
\label{eqn:pseudo_label}
\argmax_{\hat{y}_{k}} \sum_{c = 1}^{C} \hat{y}_{k}^{(c)} \log p(c;k, \theta_{h}), \ s.t. \ \hat{y}_{k} \in \Delta^{C}, \forall k,
\end{equation}
where $h$ is the category classifier that predicts $C$-category probabilities for each embedding ($e.g.$, $k$), and $\hat{y} = (\hat{y}^{(1)}, \hat{y}^{(2)}, ..., \hat{y}^{(C)})$ is the predicted category label. The key $x_{k}$ is sampled from a training dataset $X$ and encoded by the momentum encoder $f_{k}$ to get the encoded key $k = f_{k}(x_{k})$.
$\Delta^C$ denotes a probability simplex, with which a point can be represented by $C$ non-negative numbers that add up to $1$.

\end{Definition}

\begin{Remark}
\label{rm-on-CaCo-definition}
It is worth highlighting that Eq.~\ref{CaCo-Dict} only shows one group of categorical keys for the simplicity of illustration and theoretic proof. In practice, we take the same strategy as \cite{he2020momentum} and maintain a dynamic categorical dictionary with $M$-size queue ($i.e.$, $\{k^{c}_{m}\}_{1 \leq c \leq C, 1 \leq m \leq M}$), where the categorical keys are progressively updated in a category-wise manner. Specifically, for the queue of each category, we have $\{k^{c}_{1}, k^{c}_{2}, ..., k^{c}_{M}\}$, in which the oldest key is dequeued and the currently sampled key (belongs to $c$-th semantic category) is enqueued.
\end{Remark}

\textbf{Category contrastive loss.} Given the categorical dictionary $\tK = \{k^{c}_{m}\}_{1 \leq c \leq C, 1 \leq m \leq M}$ defined in Definition.~\ref{CaCo-definition}, the proposed CaCo performs contrastive learning on unlabeled target data $X_{t}$ via a category contrastive loss CatNCE that is defined by:
\begin{equation}
\label{eqn:Cat_NCE}
\begin{split}
\mathcal{L}_\mathrm{CatNCE} = &\sum_{x_{q} \in X_{t}}  -\bigg(\frac{1}{M}\sum_{m=1}^M \log \\ & \frac{\sum_{c=1}^{C} \exp(q \cdot k_{m}^{c}/\tau^{c}_{m}) (\hat{y}_{q} \times \hat{y}_{k_{m}^{c}})}{\sum_{c=1}^{C} \exp(q\cdot k_{m}^{c}/\tau^{c}_{m})} \bigg),
\end{split}
\end{equation}
where $q=f_{q}(x_q)$, $(\hat{y}_{q} \times \hat{y}_{k_{m}^{c}})$ is equal to 1 if both refer to the same category and 0 otherwise, $\tau^{c}_{m}$ is a temperature hyper-parameter and the $\cdot$ denotes the inner (dot) product. For each group of categorical keys $\{k^{1}_{m}, k^{2}_{m}, ..., k^{C}_{m}\}$, only one key is positive for the current query $q$ ($i.e.$, $(\hat{y}_{q} \times \hat{y}_{k_{m}^{c}}) = 1$) as every sample belongs to a single category. This loss is thus the log loss of a $C$-way softmax-based classifier that strives to classify $q$ as the positive key (of same category).

\begin{Remark}
\label{rm-on-Cat_NCE}
Note that the CatNCE loss in Eq.\ref{eqn:Cat_NCE} has a similar form as the InfoNCE loss in Eq.\ref{eqn:infoNCE}.
Therefore, 
InfoNCE can be interpreted as a special case of CatNCE,
where each instance (with its augmentations) itself is a category and the temperature is fixed ($i.e.$, $\tau^{c}_{m} = \tau, \forall c,m$). For CaCo, we assign different temperatures to different keys as their predicted labels have different uncertainties, $i.e.$, scaled by the prediction entropy $\mathcal{H}(\cdot)$.
The adjustable temperature parameter has also been explored in \cite{gunel2020supervised,khosla2020supervised,chen2020simple}.
\end{Remark}

\begin{Remark}
\label{rm-on-CaCo}
Note that our category contrastive loss serves as an unsupervised objective function for training the encoder networks that represent the queries and keys \cite{hadsell2006dimensionality}. In general, the query representation is $q=f_\textrm{q}(x^q)$ where $f_\textrm{q}$ is an encoder network and $x^q$ is a query sample (likewise, $k=f_\textrm{k}(x^k)$). 
Their instantiations depend on the specific pretext task.
The input $x^q$ and $x^k$ can be images \cite{hadsell2006dimensionality,wu2018unsupervised,ye2019unsupervised}, patches \cite{oord2018representation} or context consisting of a set of patches \cite{oord2018representation}, etc.
The networks $f_\textrm{q}$ and $f_\textrm{k}$ can be identical \cite{hadsell2006dimensionality,wang2015unsupervised,ye2019unsupervised}, partially shared \cite{oord2018representation,bachman2019learning}, or different \cite{tian2019contrastive,he2020momentum}.
\end{Remark}

\textbf{Relations to existing instance contrast methods}. Beyond instance-discriminative representations as learnt by instance contrast~\cite{wu2018unsupervised,ye2019unsupervised, he2020momentum, oord2018representation, tian2019contrastive,chen2020simple}, CaCo learns category-discriminative yet domain-invariant representation.

\subsection{Theoretical Insights}
The category contrast (CaCo) is inherently connected with some probabilistic models. Specifically, CaCo can be modeled as an example of Expectation Maximization (EM):

\begin{Proposition} The category contrastive learning can be modeled as a maximum likelihood (ML) problem optimized via Expectation Maximization (EM).
\end{Proposition}

\begin{Proposition} The categorical contrastive learning is convergent under certain conditions.
\end{Proposition}

The proofs of \textbf{Propositions 1} and \textbf{2} are provided in the Appendix.

\section{Experiments}

This section presents experimental results.
Sections \ref{exp_dataset} and \ref{exp_impl} describe the dataset and implementation details. Sections \ref{exp_uda_seg}, \ref{exp_uda_obj} and \ref{exp_uda_cla} present the UDA experiments in segmentation, detection and classification, respectively. Section \ref{exp_discussion} discusses different features of the proposed method.

\renewcommand\arraystretch{1.2}
\begin{table*}[!t]
	\centering
	\resizebox{\linewidth}{!}{
	\begin{tabular}{c|ccccccccccccccccccc|c}
		\hline
		Method  & Road & SW & Build & Wall & Fence & Pole & TL & TS & Veg. & Terrain & Sky & PR & Rider & Car & Truck & Bus & Train & Motor & Bike & mIoU\\
		\hline
		Baseline \cite{chen2017deeplab} &75.8	&16.8	&77.2	&12.5	&21.0	&25.5	&30.1	&20.1	&81.3	&24.6	&70.3	&53.8	&26.4	&49.9	&17.2	&25.9	&6.5	&25.3	&36.0	&36.6\\\hline
		\textbf{CaCo-S}  &91.1	&54.4	&79.6	&27.0	&22.9	&36.9	&40.2	&33.4	&83.7	&36.3	&65.2	&59.7	&22.4	&83.5	&37.5	&49.3	&10.1	&23.3	&31.8	&46.8
		\\
		\textbf{CaCo-T}  &92.0	&53.5	&81.6	&28.9	&26.3	&36.5	&42.7	&36.3	&81.8	&37.2	&75.5	&59.8	&26.5	&84.9	&40.0	&44.9	&11.6	&27.0	&29.9	&48.3
		\\
		\textbf{CaCo} &91.9	&54.3	&82.7	&31.7	&25.0	&38.1	&46.7	&39.2	&82.6	&39.7	&76.2	&63.5	&23.6	&85.1	&38.6	&47.8	&10.3	&23.4	&35.1	&49.2
		\\\hline
		AdaptSeg~\cite{tsai2018learning}  &86.5	&36.0	&79.9	&23.4	&23.3	&23.9	&35.2	&14.8	&83.4	&33.3	&75.6	&58.5	&27.6	&73.7	&32.5	&35.4	&3.9	&30.1	&28.1	&42.4\\
		CBST~\cite{zou2018unsupervised} &91.8 &53.5 &80.5 &32.7 &21.0 &34.0 &28.9 &20.4 &83.9 &34.2 &80.9 &53.1 &24.0 &82.7 &30.3 &35.9 &16.0 &25.9 &42.8 &45.9\\
        CLAN~\cite{luo2019taking}  &87.0 &27.1 &79.6 &27.3 &23.3 &28.3 &35.5 &24.2 &83.6 &27.4 &74.2 &58.6 &28.0 &76.2 &33.1 &36.7 &{6.7} &{31.9} &31.4 &43.2\\
        AdvEnt~\cite{vu2019advent}  &89.4 &33.1 &81.0 &26.6 &26.8 &27.2 &33.5 &24.7 &{83.9} &{36.7} &78.8 &58.7 &30.5 &{84.8} &38.5 &44.5 &1.7 &31.6 &32.4 &45.5\\
        IDA~\cite{pan2020unsupervised}  &90.6 &37.1 &82.6 &30.1 &19.1 &29.5 &32.4 &20.6 &85.7 &40.5 &79.7 &58.7 &31.1 &86.3 &31.5 &48.3 &0.0 &30.2 &35.8 &46.3\\
        BDL~\cite{li2019bidirectional}  &91.0  &44.7  &84.2  &34.6  &27.6 &30.2  &36.0  &36.0 &85.0  &43.6  &83.0 &58.6 &31.6  &83.3  &35.3  &49.7 &3.3  &28.8  &35.6 &48.5\\
        CrCDA~\cite{huang2020contextual}  &92.4	&55.3	&82.3	&31.2    &29.1	&32.5	&33.2	&35.6	&83.5	&34.8	&84.2	&58.9	&32.2	&84.7	&40.6	&46.1	&2.1	&31.1	&32.7	&48.6 \\
        SIM~\cite{wang2020differential}  &90.6 &44.7 &84.8 &34.3 &28.7 &31.6 &35.0 &37.6 &84.7 &43.3 &85.3 &57.0 &31.5 &83.8 &42.6 &48.5 &1.9 &30.4 &39.0 &49.2\\
        TIR~\cite{kim2020learning}  &92.9 &55.0 &85.3 &34.2 &31.1 &34.9 &40.7 &34.0 &85.2 &40.1 &87.1 &61.0 &31.1 &82.5 &32.3 &42.9 &0.3 &36.4 &46.1 &50.2\\
        \hline
        CRST~\cite{zou2019confidence}  &91.0 &55.4 &80.0 &33.7 &21.4 &37.3 &32.9 &24.5 &85.0 &34.1 &80.8 &57.7 &24.6 &84.1 &27.8 &30.1 &26.9 &26.0 &42.3 &47.1\\
        \textbf{+CaCo}  &93.0	&58.4	&83.1	&34.0	&29.3	&37.0	&47.1	&42.9	&84.6	&41.5	&82.8	&61.8	&32.2	&86.9	&39.2	&48.0	&22.4	&31.1	&45.7	&52.7
        \\\hline
        FDA~\cite{yang2020fda}  &92.5 &53.3 &82.4 &26.5 &27.6 &36.4 &40.6 &38.9 &82.3 &39.8 &78.0 &62.6 &34.4 &84.9 &34.1 &53.1 &16.9 &27.7 &46.4 &50.5\\
        \textbf{+CaCo}  &93.2	&54.5	&84.6	&32.9	&29.3	&39.7	&46.9	&42.7	&84.4	&40.1	&83.7	&61.1	&32.2	&85.6	&41.7	&51.2	&19.2	&35.6	&45.9	&52.9
        \\\hline
        ProDA~\cite{zhang2021proda} &87.8& 56.0& 79.7 &46.3& 44.8& 45.6& 53.5& 53.5& 88.6& 45.2& 82.1& 70.7& 39.2& 88.8& 45.5& 59.4& 1.0& 48.9& 56.4& 57.5\\
\textbf{+CaCo} &93.8	&64.1	&85.7	&43.7	&42.2	&46.1	&50.1	&54.0	&88.7	&47.0	&86.5	&68.1	&2.9	&88.0	&43.4	&60.1	&31.5	&46.1	&60.9	&58.0\\
\hline
	\end{tabular}
	}
	\caption{
	Experiments over UDA-based semantic segmentation task GTA5 $\rightarrow$ Cityscapes:  CaCo-S, CaCo-T and CaCo construct the category-aware dictionary by sampling key samples $x_{k}$ from the source dataset $X_{s}$ only, the target dataset $X_{t}$ only, and both datasets, respectively.}
	\label{table:gta2city}
\end{table*}

\renewcommand\arraystretch{1.2}
\begin{table*}[h]
	\centering
	\resizebox{0.9\linewidth}{!}{
	\begin{tabular}{c|cccccccccccccccc|c|c}
		\hline
		Method  & Road & SW & Build & Wall\textsuperscript{*} & Fence\textsuperscript{*} & Pole\textsuperscript{*} & TL & TS & Veg. & Sky & PR & Rider & Car & Bus & Motor & Bike & mIoU & mIoU\textsuperscript{*}\\
		\hline
		Baseline \cite{chen2017deeplab} &55.6	&23.8	&74.6	&9.2	&0.2	&24.4	&6.1	&12.1	&74.8	&79.0	&55.3	&19.1	&39.6	&23.3	&13.7	&25.0	&33.5	&38.6\\ \hline
		PatAlign~\cite{tsai2019domain}  &82.4 &38.0 &78.6 &8.7 &0.6 &26.0 &3.9 &11.1 &75.5 &84.6 &53.5 &21.6 &71.4 &32.6 &19.3 &31.7 &40.0 &46.5\\
        AdaptSeg~\cite{tsai2018learning}  &84.3 &42.7 &77.5 &- &- &- &4.7 &7.0 &77.9 &82.5 &54.3 &21.0 &72.3 &32.2 &18.9 &32.3 &- &46.7\\
        CLAN~\cite{luo2019taking}  &81.3 &37.0 &{80.1} &- &- &- &{16.1} &{13.7} &78.2 &81.5 &53.4 &21.2 &73.0 &32.9 &{22.6} &30.7 &- &47.8\\
        AdvEnt~\cite{vu2019advent}  &85.6 &42.2 &79.7 &{8.7} &0.4 &25.9 &5.4 &8.1 &{80.4} &84.1 &{57.9} &23.8 &73.3 &36.4 &14.2 &{33.0} &41.2 &48.0\\
        IDA~\cite{pan2020unsupervised}  &84.3 &37.7 &79.5 &5.3 &0.4 &24.9 &9.2 &8.4 &80.0 &84.1 &57.2 &23.0 &78.0 &38.1 &20.3 &36.5 &41.7 &48.9\\
        CrCDA~\cite{huang2020contextual}   &86.2	&44.9	&79.5	&8.3	&0.7	&27.8	&9.4	&11.8	&78.6	&86.5	&57.2	&26.1	&76.8	&39.9	&21.5	&32.1	&42.9	&50.0\\
        TIR~\cite{kim2020learning}  &92.6 &53.2 &79.2 &- &- &- &1.6 &7.5 &78.6 &84.4 &52.6 &20.0 &82.1 &34.8 &14.6 &39.4 &- &49.3\\
        SIM~\cite{wang2020differential}  &83.0 &44.0 &80.3 &- &- &- &17.1 &15.8 &80.5 &81.8 &59.9 &33.1 &70.2 &37.3 &28.5 &45.8 &- &52.1\\
        BDL~\cite{li2019bidirectional}  &86.0   &46.7   &80.3&-&-&-  &14.1   &11.6 &79.2 &81.3 &54.1   &27.9   &73.7   &42.2   &25.7   &45.3  &- &51.4\\
        \hline
        CRST~\cite{zou2019confidence}  &67.7 &32.2 &73.9 &10.7 &1.6 &37.4 &22.2 &31.2 &80.8 &80.5 &60.8 &29.1 &82.8 &25.0 &19.4 &45.3 &43.8 &50.1\\
        \textbf{+CaCo} &88.8	&48.0	&79.5	&6.9	&0.3	&36.9	&28.0	&22.1	&83.5	&84.1	&63.9	&31.0	&85.8	&38.1	&29.4	&49.1	&48.5	&56.2
        \\\hline
        FDA~\cite{yang2020fda}  &79.3 &35.0 &73.2 &- &- &- &19.9 &24.0 &61.7 &82.6 &61.4 &31.W1 &83.9 &40.8 &38.4 &51.1 &- &52.5\\
        \textbf{+CaCo} &86.4	&43.3	&78.7	&9.0	&0.1	&28.5	&26.7	&29.7	&81.7	&82.9	&59.3	&28.1	&82.9	&38.6	&35.7	&50.0	&47.6	&55.7
        \\\hline
        \textbf{CaCo} &87.4	&48.9	&79.6	&8.8	&0.2	&30.1	&17.4	&28.3	&79.9	&81.2	&56.3	&24.2	&78.6	&39.2	&28.1	&48.3	&46.0	&53.6
        \\\hline
	\end{tabular}
	}
	\caption{
	Experiments over UDA-based semantic segmentation task SYNTHIA $\rightarrow$ Cityscapes.
	}
	\label{table:synthia2city}
\end{table*}

\subsection{Datasets}
\label{exp_dataset}

\textbf{Adaptation for semantic segmentation:} It involves three public datasets over two challenging UDA tasks, $i.e.$, GTA5 \cite{richter2016playing} $\rightarrow$ Cityscapes \cite{cordts2016cityscapes} and SYNTHIA \cite{ros2016synthia} $\rightarrow$ Cityscapes. Specifically, GTA5 is a synthesized dataset with $24,966$ images and $19$ common categories with Cityscapes. SYNTHIA is a synthesized dataset with $9,400$ images and $16$ common categories with Cityscapes. Cityscapes is a real-image dataset with $2975$ training images and $500$ validation images.

\textbf{Adaptation for object detection:} It involves three public datasets over two adaptation tasks, $i.e.$, Cityscapes $\rightarrow$ Foggy Cityscapes~\cite{sakaridis2018semantic} and Cityscapes $\rightarrow$ BDD100k~\cite{yu2018bdd100k}. Specifically, Foggy Cityscapes is a synthesized dataset that applies simulated fog on Cityscapes images. BDD100k is a real dataset with $70k$ training images, $10k$ validation images and $7$ common categories with Cityscapes. As in~\cite{xu2020exploring,saito2019strong,chen2018domain}, we only use a subset of BDD100k ``\textit{daytime}'' in experiments.

\textbf{Adaptation for image classification:} It involves two adaptation benchmarks VisDA17~\cite{peng2018visda} and Office-31~\cite{saenko2010adapting}. VisDA17 consists of a source domain with $152,409$ synthesized images of $12$ categories and a target domain with $55,400$ real images. Office-31 consists of images of $31$ categories which were collected from Amazon (2817 images), Webcam (795 images) and DSLR (498 images), respectively.
The evaluation is on every 
pair of them as in~\cite{zou2019confidence, saenko2010adapting, sankaranarayanan2018generate}.

\renewcommand\arraystretch{1.2}
\begin{table*}[t]
\centering
\resizebox{0.6\linewidth}{!}{
\begin{tabular}{cccccccccc}
\hline
Method & person & rider & car & truck & bus & train & mcycle & bicycle & mAP \\
\hline
Baseline \cite{ren2015faster} &  24.4 & 30.5  & 32.6 & 10.8  & 25.4 & 9.1 & 15.2 & 28.3 & 22.0 \\ \hline
MAF~\cite{he2019multi} &  28.4 & 39.5  & 43.9 & 23.8  & 39.9 & 33.3 & 29.2 & 33.9 & 34.0 \\
SCDA~\cite{zhu2019adapting} &  33.5 & 38.0  & 48.5 & 26.5  & 39.0 & 23.3 & 28.0 & 33.6 & 33.8 \\
DA~\cite{chen2018domain} &25.0 &31.0 &40.5 &22.1 &35.3 &20.2 &20.0 &27.1 &27.6\\
MLDA~\cite{Xie_2019_ICCV} &33.2 &44.2 &44.8 &28.2  &41.8  &28.7 &30.5 &36.5 &36.0\\
DMA~\cite{kim2019diversify} &30.8 &40.5 &44.3 &27.2 &38.4 &34.5 &28.4 &32.2 &34.6\\
CAFA~\cite{hsu2020every} &41.9 &38.7 &56.7 &22.6 &41.5 &26.8 &24.6 &35.5 &36.0 \\
\hline
SWDA~\cite{saito2019strong} &36.2 &35.3   &43.5 &30.0 &29.9 &42.3  &32.6 &24.5 &34.3\\
\textbf{+CaCo} &39.3	&46.1	&48.0	&32.4	&45.7	&38.7	&31.3	&35.3	&39.6
\\\hline
CRDA~\cite{xu2020exploring} & 32.9 & 43.8  & 49.2 & 27.2  & 45.1 & 36.4 & 30.3 & 34.6 & 37.4 \\
\textbf{+CaCo} &39.4	&47.4	&47.9	&32.5	&46.4	&39.9	&32.7	&35.4	&40.2
\\\hline
\textbf{CaCo} &38.3	&46.7	&48.1	&33.2	&45.9	&37.6	&31.0	&33.0	&39.2
\\\hline
\end{tabular}
}
\caption{Experiments over UDA-based object detection task Cityscapes $\rightarrow$ Foggy Cityscapes.}
\label{table:det_city2fog}
\end{table*}

\renewcommand\arraystretch{1.2}
\begin{table*}[h]
\centering
\resizebox{0.56\linewidth}{!}{
\begin{tabular}{ccccccccc}
\hline
Method & person & rider & car & truck & bus & mcycle & bicycle & mAP \\
\hline
Baseline \cite{ren2015faster} &  26.9 & 22.1 & 44.7 & 17.4  & 16.7 & 17.1 & 18.8 & 23.4 \\ 
DA~\cite{chen2018domain} & 29.4 & 26.5  & 44.6 & 14.3  & 16.8 & 15.8 & 20.6 & 24.0 \\
\hline
SWDA~\cite{saito2019strong} & 30.2 & 29.5  & 45.7 & 15.2  & 18.4 & 17.1 & 21.2 & 25.3 \\
\textbf{+CaCo} &32.1	&32.9	&51.6	&20.5	&23.7	&20.1	&25.6	&29.5
\\\hline
CRDA~\cite{xu2020exploring} & 31.4 & 31.3  & 46.3 & 19.5  & 18.9 & 17.3 & 23.8 & 26.9 \\
\textbf{+CaCo} &32.5	&34.1	&51.1	&21.6	&25.1	&20.5	&26.5	&30.2
\\\hline
\textbf{CaCo} &32.7	&32.2	&50.6	&20.2	&23.5	&19.4	&25.0	&29.1
\\\hline
\end{tabular}
}
\caption{
Experiments over UDA-based object detection tasks Cityscapes $\rightarrow$ BDD100k.
}
\label{table:det_city2BDD}
\end{table*}

\subsection{Implementation Details}
\label{exp_impl}
\textbf{Semantic segmentation:}
As in~\cite{tsai2018learning,zou2018unsupervised}, we utilize DeepLab-V2~\cite{chen2017deeplab} with ResNet101 \cite{he2016deep} as the segmentation backbone. We employ SGD~\cite{bottou2010large} as the optimizer with momentum $0.9$, weight decay $1e-4$ and learning rate $2.5e-4$. The learning rate is decayed by a polynomial annealing policy~\cite{chen2017deeplab}.

\textbf{Object detection:} Following~\cite{xu2020exploring, saito2019strong, chen2018domain}, we employ Faster R-CNN~\cite{ren2015faster} with VGG-16~\cite{simonyan2014very} as the detection backbone. We adopt SGD optimizer \cite{bottou2010large} with momentum $0.9$ and weight decay $5e-4$. The learning rate is $1e-3$ for first $50k$ iterations and then decreased to $1e-4$ for $20k$ iterations \cite{xu2020exploring,saito2019strong,chen2018domain}. The image shorter side is set to 600 and RoIAlign is employed for feature extraction.

\textbf{Image classification:} Following ~\cite{zou2019confidence, saenko2010adapting,sankaranarayanan2018generate}, we employ ResNet-101 and ResNet-50~\cite{he2016deep} as the classification backbones for VisDA17 and Office-31, respectively. We adopt SGD as the optimizer \cite{bottou2010large} with momentum $0.9$, weight decay $5e-4$, learning rate $1e-3$ and batch size $32$.

We set the length of dictionary queue $M$ at $100$ in all experiments except in parameter analysis. In addition, we set the momentum update coefficient $b$ at $0.999$ and the basic temperature $\tau$ at $0.07$ as in \cite{he2020momentum}.

\subsection{UDA for Semantic Segmentation}
\label{exp_uda_seg}

\begin{figure*}[h]
\begin{tabular}{p{3.66cm}p{3.66cm}p{3.66cm}p{3.66cm}}
& \raisebox{-0.5\height}{\includegraphics[width=1.1\linewidth,height=0.55\linewidth]{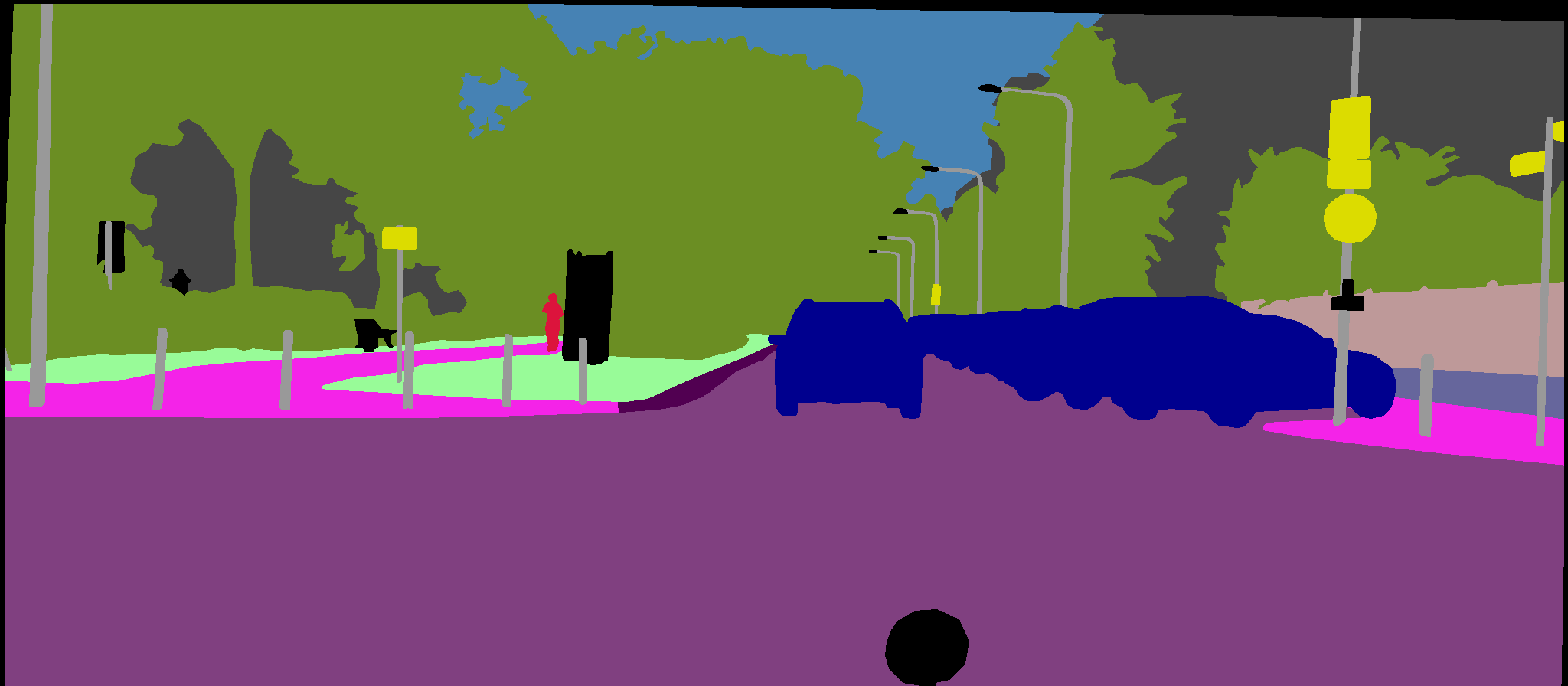}}
& \raisebox{-0.5\height}{\includegraphics[width=1.1\linewidth,height=0.55\linewidth]{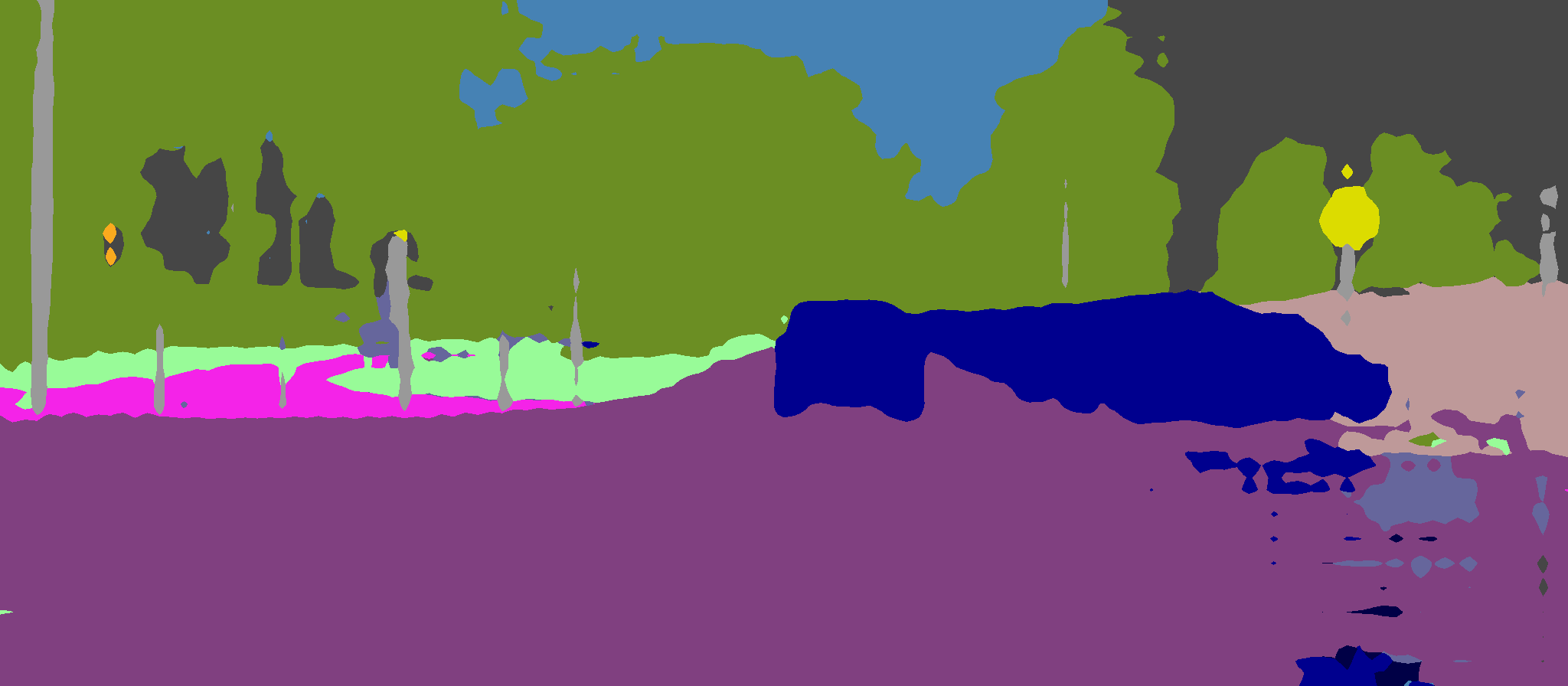}}
& \raisebox{-0.5\height}{\includegraphics[width=1.1\linewidth,height=0.55\linewidth]{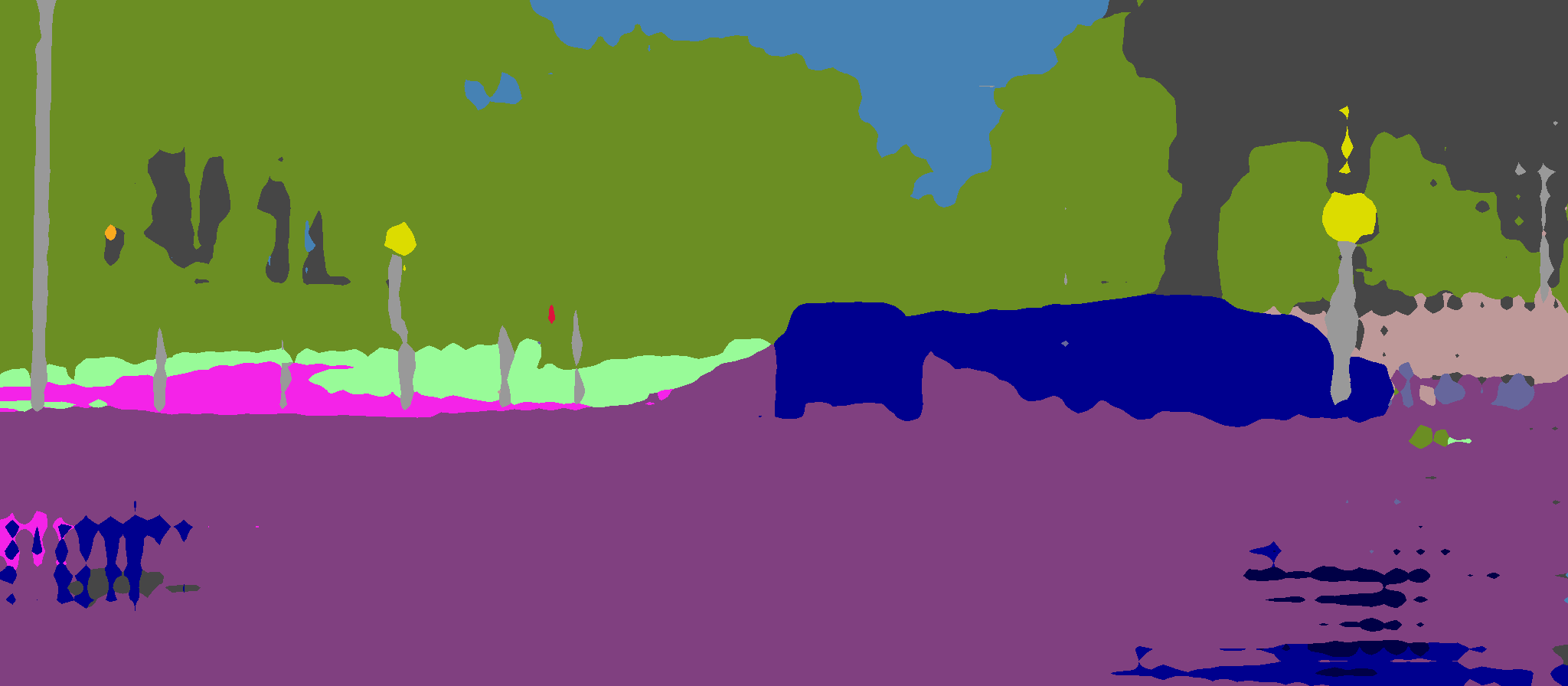}}
\\
& \raisebox{-0.5\height}{Ground Truth}
& \raisebox{-0.5\height}{ADVENT}
& \raisebox{-0.5\height}{CRST}\vspace{0.5mm}
\\
\raisebox{-0.5\height}{\includegraphics[width=1.1\linewidth,height=0.55\linewidth]{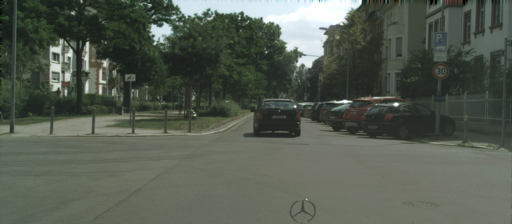}}
& \raisebox{-0.5\height}{\includegraphics[width=1.1\linewidth,height=0.55\linewidth]{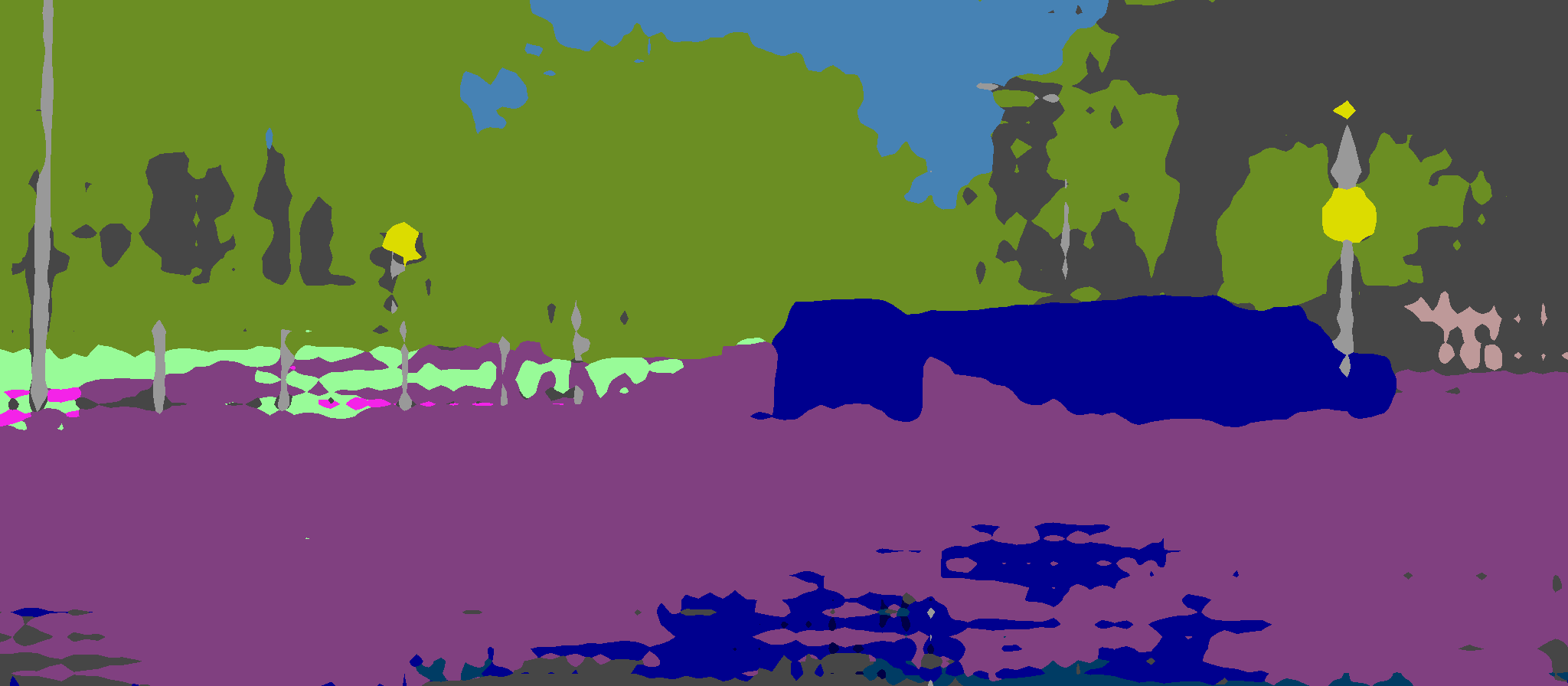}}
& \raisebox{-0.5\height}{\includegraphics[width=1.1\linewidth,height=0.55\linewidth]{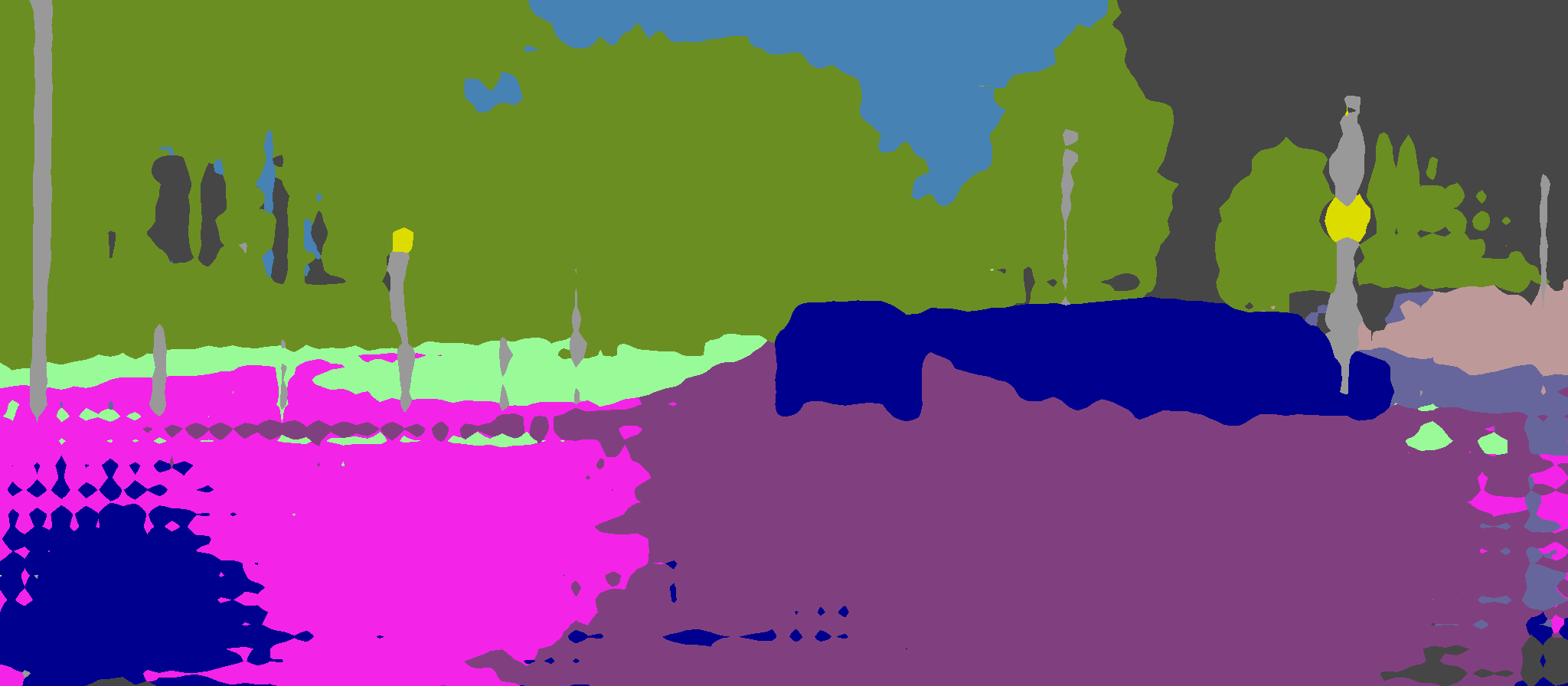}}
& \raisebox{-0.5\height}{\includegraphics[width=1.1\linewidth,height=0.55\linewidth]{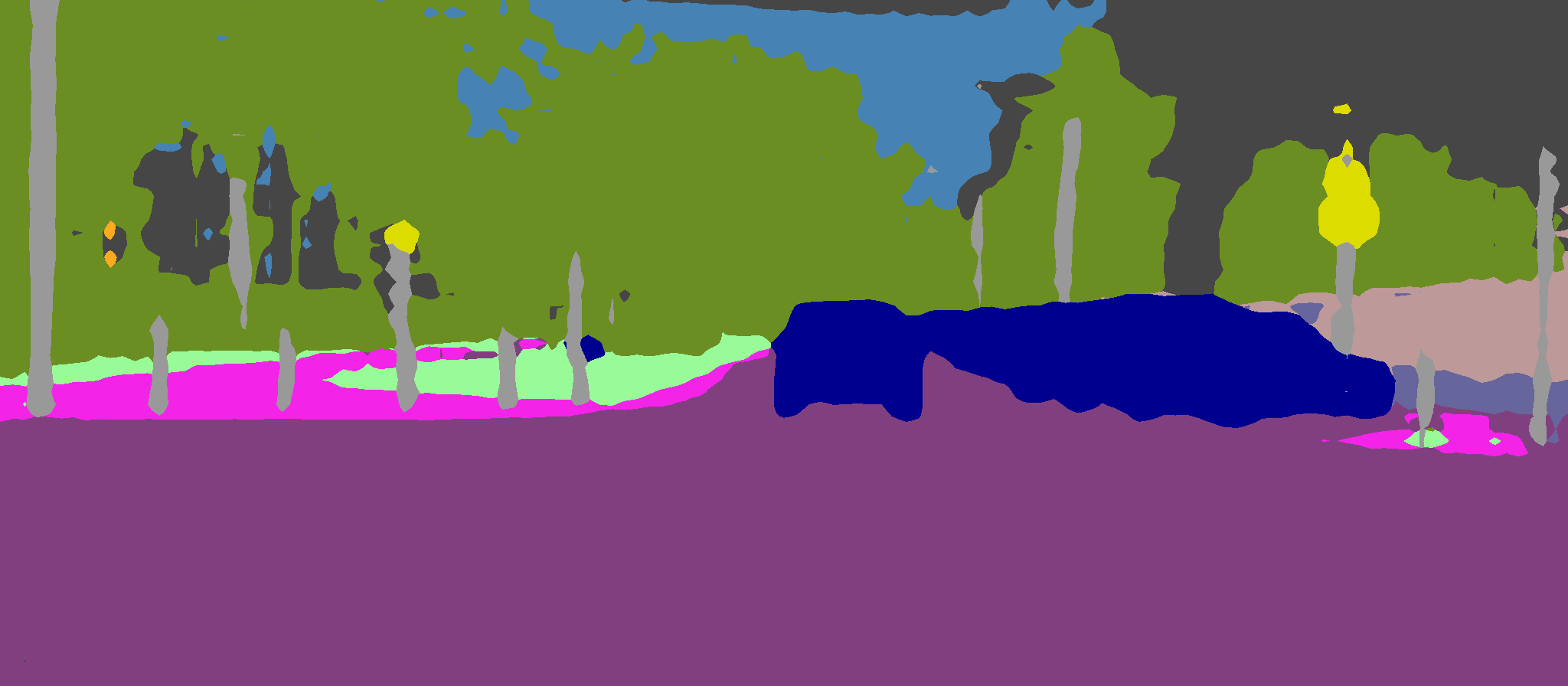}}
\\
\raisebox{-0.5\height}{Target Image}
& \raisebox{-0.5\height}{Baseline}
& \raisebox{-0.5\height}{FDA}
& \raisebox{-0.5\height}{\textbf{CaCo(Ours)}}\vspace{1mm}
\\
& \raisebox{-0.5\height}{\includegraphics[width=1.1\linewidth,height=0.55\linewidth]{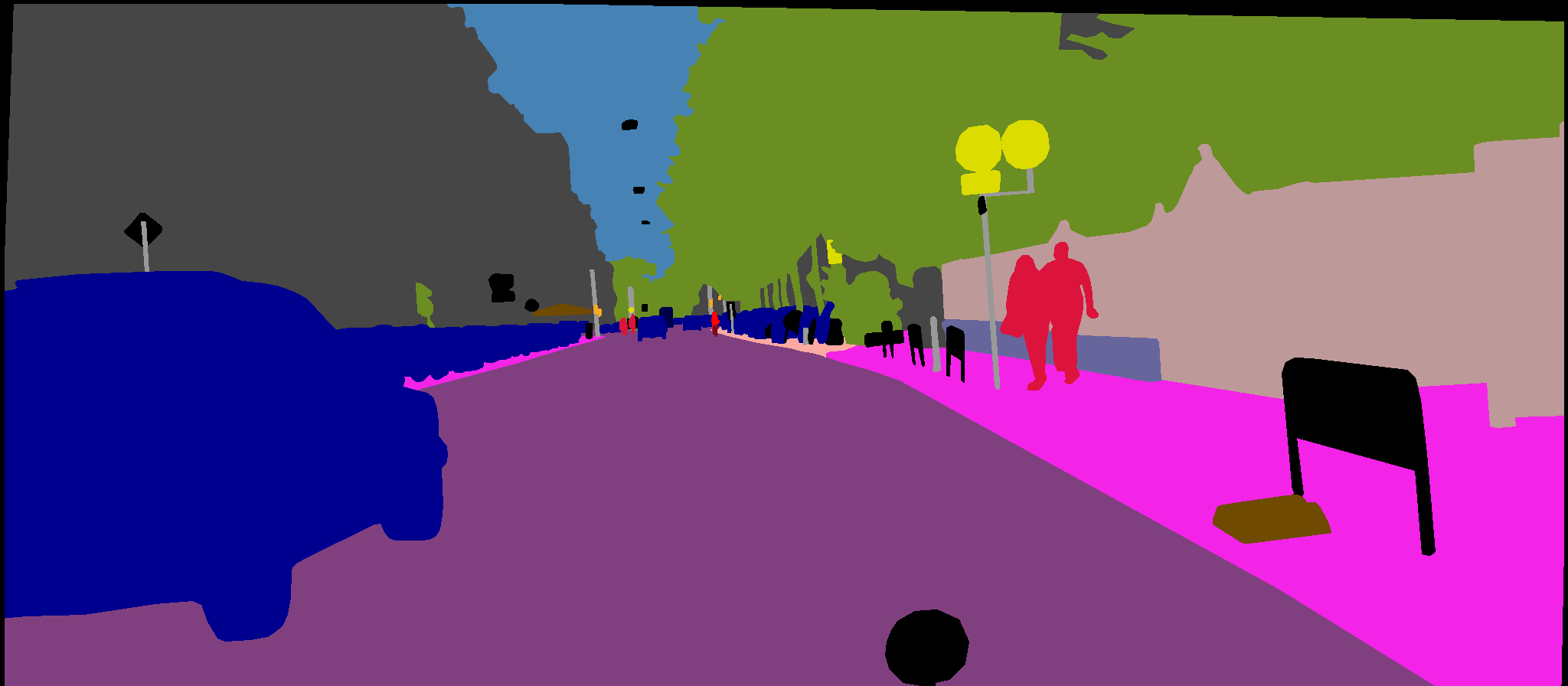}}
& \raisebox{-0.5\height}{\includegraphics[width=1.1\linewidth,height=0.55\linewidth]{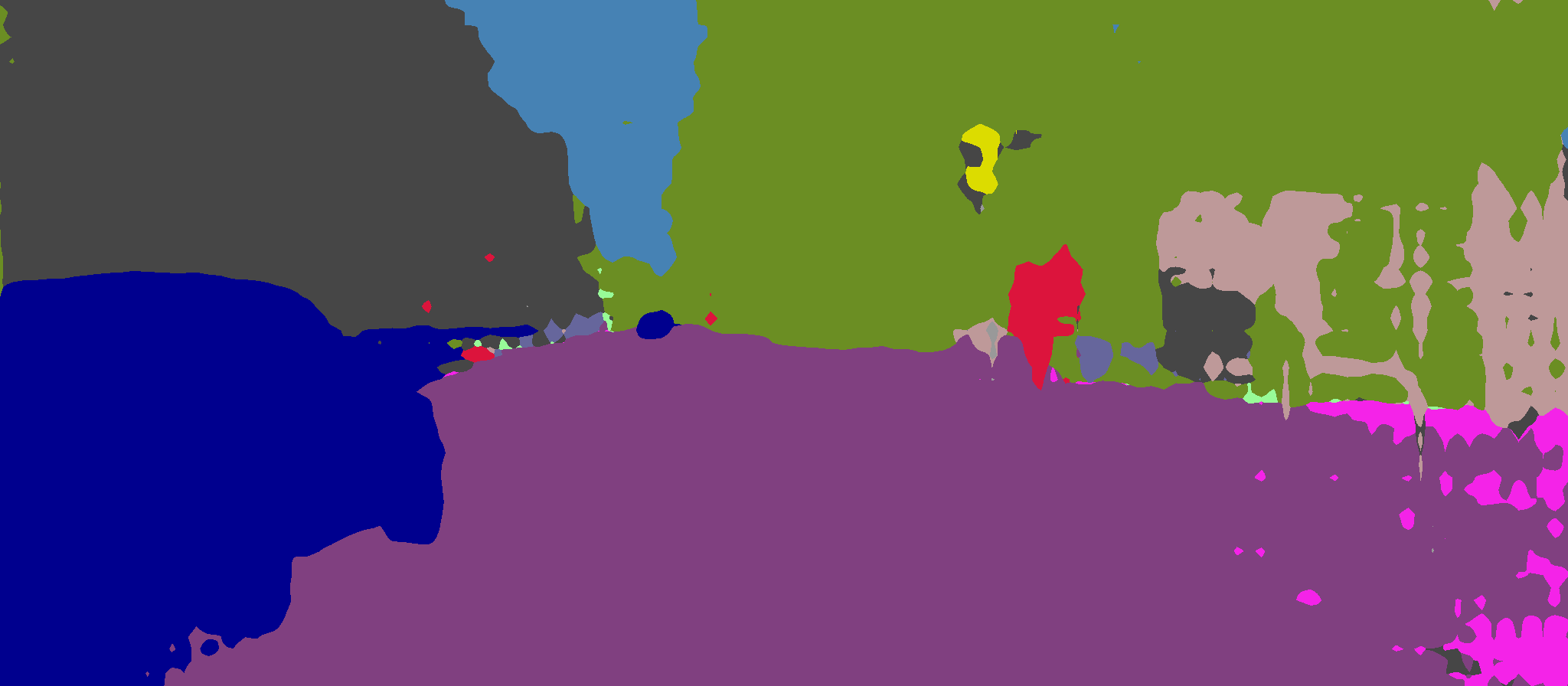}}
& \raisebox{-0.5\height}{\includegraphics[width=1.1\linewidth,height=0.55\linewidth]{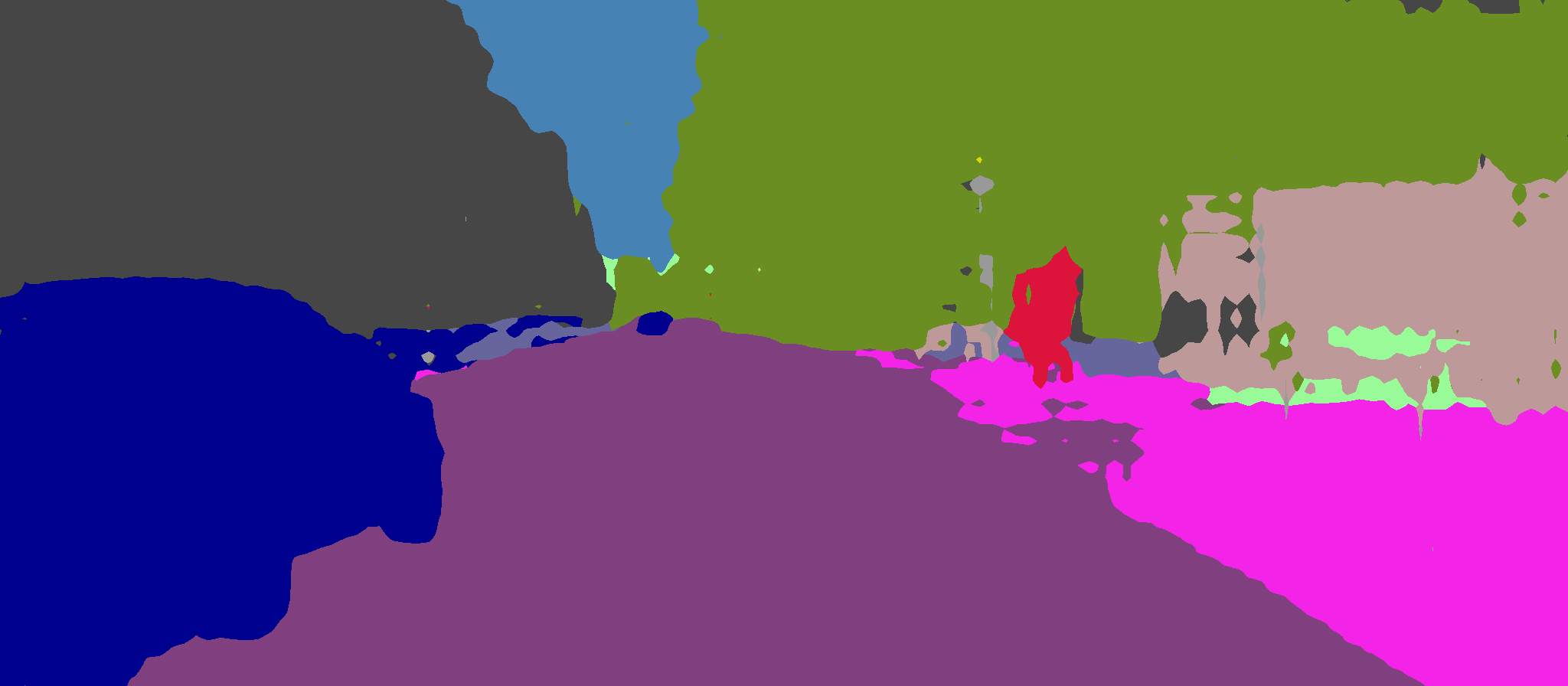}}
\\
 & \raisebox{-0.5\height}{Ground Truth}
& \raisebox{-0.5\height}{ADVENT}
& \raisebox{-0.5\height}{CRST}\vspace{0.5mm}
\\
\raisebox{-0.5\height}{\includegraphics[width=1.1\linewidth,height=0.55\linewidth]{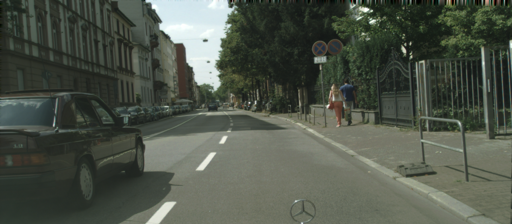}}
&\raisebox{-0.5\height}{\includegraphics[width=1.1\linewidth,height=0.55\linewidth]{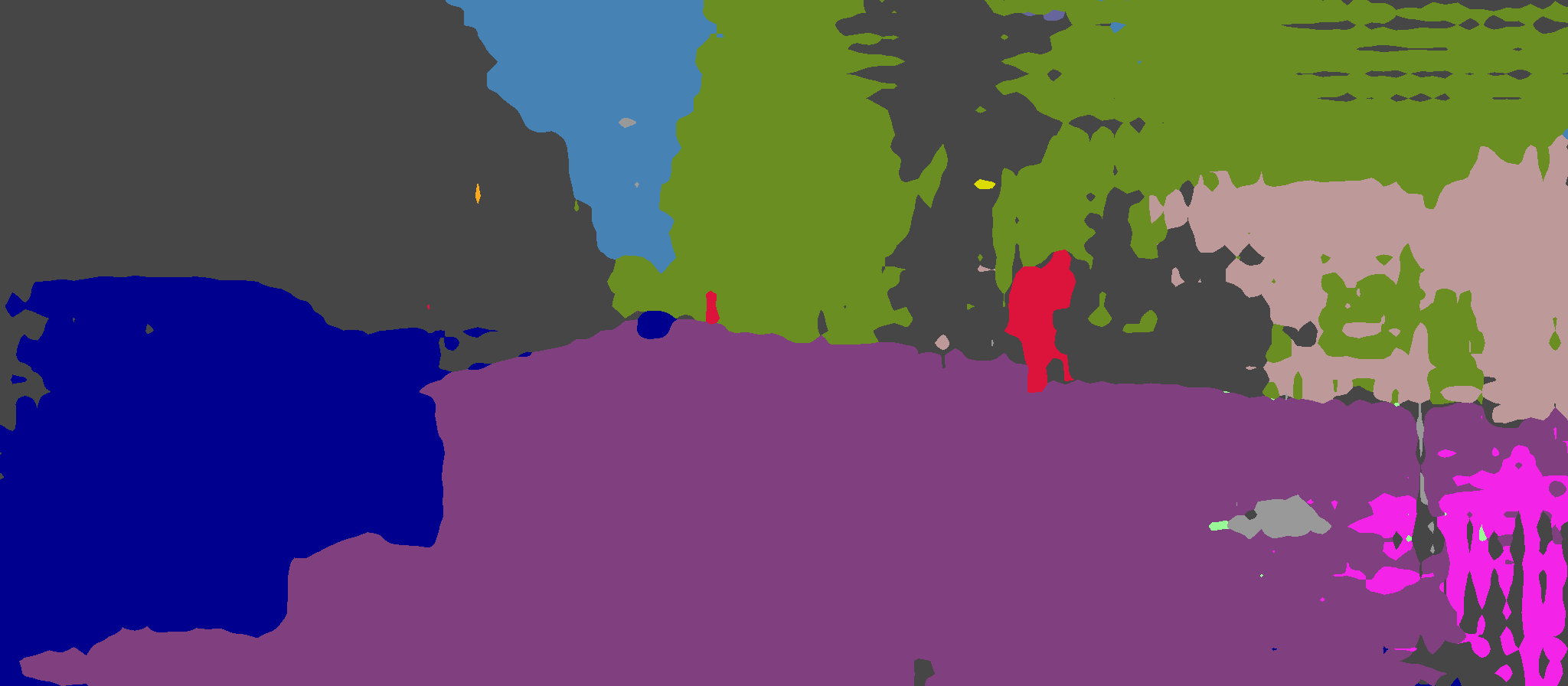}}
& \raisebox{-0.5\height}{\includegraphics[width=1.1\linewidth,height=0.55\linewidth]{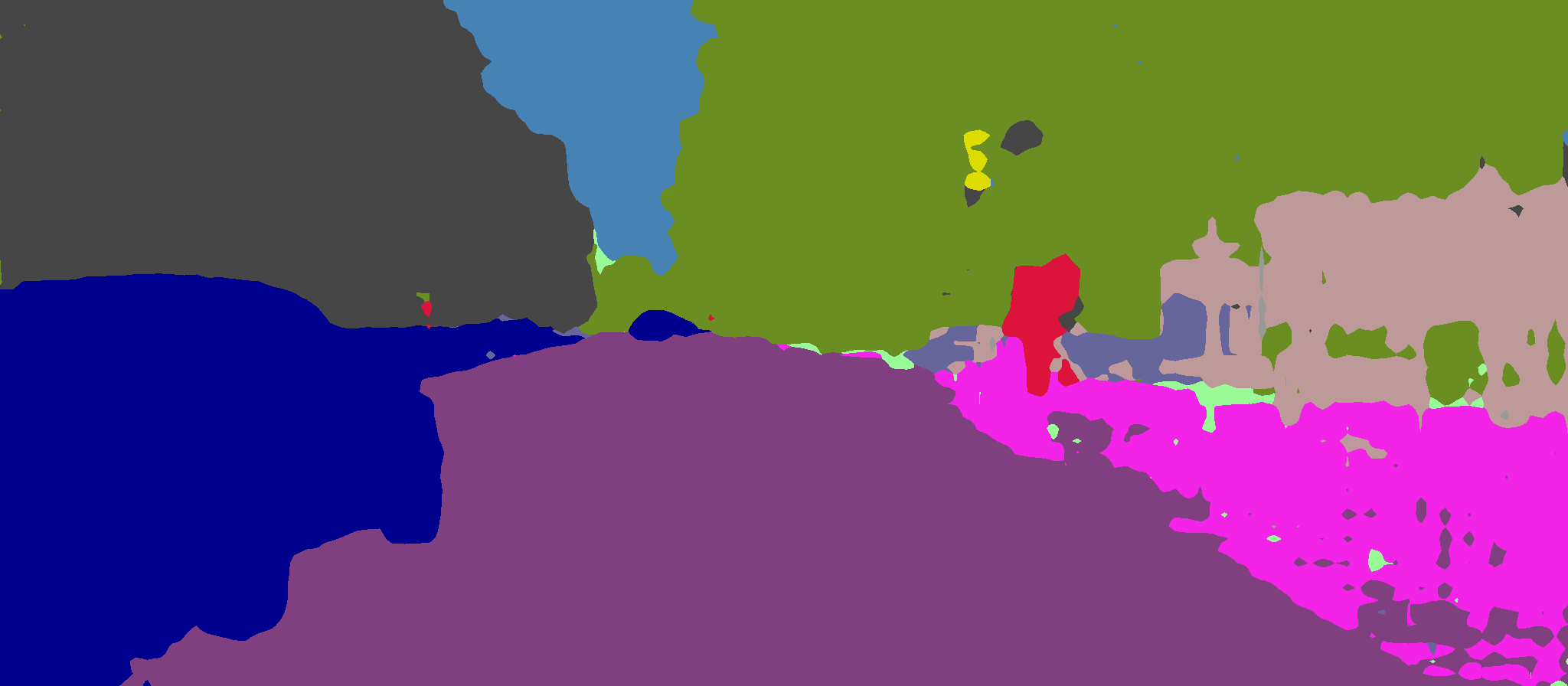}}
& \raisebox{-0.5\height}{\includegraphics[width=1.1\linewidth,height=0.55\linewidth]{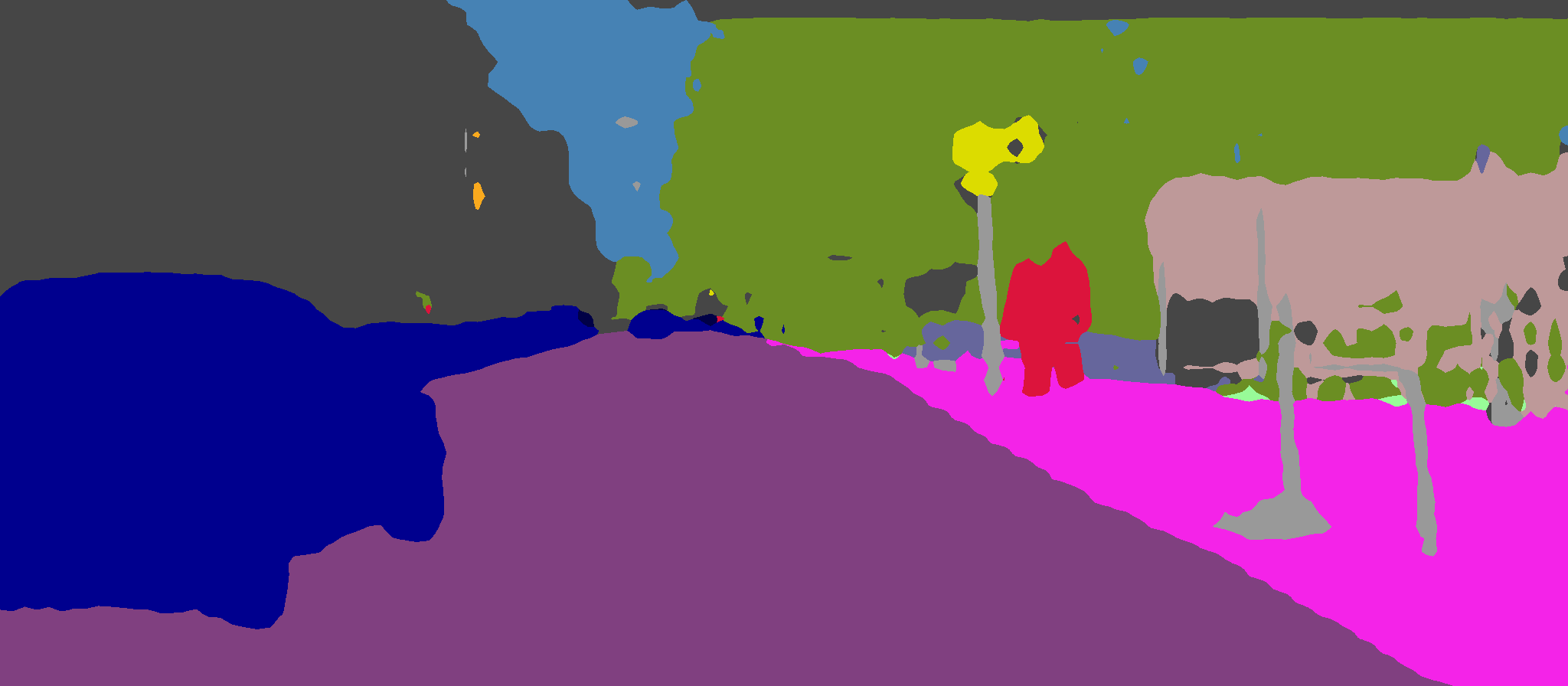}}
\\
\raisebox{-0.5\height}{Target Image}
&\raisebox{-0.5\height}{Baseline}
& \raisebox{-0.5\height}{FDA}
& \raisebox{-0.5\height}{\textbf{CaCo(Ours)}}\vspace{1mm}
\\
\end{tabular}
\caption{
Qualitative comparisons over domain adaptive semantic segmentation task GTA5 $\rightarrow$ Cityscapes.
}
\label{fig:results}
\end{figure*}

Table~\ref{table:gta2city} reports semantic segmentation results on the task GTA5 $\rightarrow$ Cityscapes. It can be seen that the proposed CaCo achieves comparable performance with 
state-of-the-art methods.
In addition, CaCo is complementary to existing UDA approaches that exploit adversarial loss, image translation loss and self-training loss.
As shown in Table~\ref{table:gta2city}, incorporating CaCo as denoted by ``+CaCo'' boosts the performance of state-of-the-art methods clearly and consistently. Fig.~\ref{fig:results} presents the qualitative comparisons.

\textbf{Ablation studies.} We perform ablation studies over a widely adopted \textit{Baseline}~\cite{he2016deep} as shown on the top of Table~\ref{table:gta2city}, where \textit{CaCo-S}, \textit{CaCo-T} and \textit{CaCo} mean that the category-aware dictionary is built with keys from source domain, target domain and both, respectively. It can be seen that \textit{CaCo-S} and \textit{CaCo-T} both outperform the \textit{Baseline} by large margins. \textit{CaCo-S} and \textit{CaCo-T} provide orthogonal self-supervision signals, where \textit{CaCo-S} focuses on inter-domain category contrastive learning between target samples and source keys and \textit{CaCo-T} focuses on intra-domain category contrastive learning between target samples and target keys. In addition, \textit{CaCo} performs clearly the best, showing that the keys from the source and target domains are complementary.

\renewcommand\arraystretch{1.2}
\begin{table*}[ht]
\centering
\resizebox{0.92\linewidth}{!}{
\begin{tabular}{c|cccccccccccc|c}
		\hline
		Method & Aero & Bike & Bus & Car & Horse & Knife & Motor & Person & Plant & Skateboard & Train & Truck & Mean\\
		\hline
		Baseline \cite{he2016deep} & 55.1 & 53.3 & 61.9 & 59.1 & 80.6 & 17.9 & 79.7 & 31.2 & 81.0 & 26.5 & 73.5 & 8.5 & 52.4\\ \hline
		MMD \cite{long2015learning} & 87.1 & 63.0 & 76.5 & 42.0 & 90.3 & 42.9 & 85.9 & 53.1 & 49.7 & 36.3 & 85.8 & 20.7 & 61.1\\
		DANN \cite{ganin2016domain} & 81.9 & 77.7 & 82.8 & 44.3 & 81.2 & 29.5 & 65.1 & 28.6 & 51.9 & 54.6 & 82.8 & 7.8 & 57.4\\ 
		ENT \cite{grandvalet2005semi} & 80.3 & 75.5 & 75.8 & 48.3 & 77.9 & 27.3 & 69.7 & 40.2 & 46.5 & 46.6 & 79.3 & 16.0 & 57.0\\
		MCD \cite{saito2018maximum} & 87.0 & 60.9 & {83.7} & 64.0 & 88.9 & 79.6 & 84.7 & {76.9} & {88.6} & 40.3 & 83.0 & 25.8 & 71.9\\
		ADR \cite{saito2018adversarial} & 87.8 & 79.5 & {83.7} & 65.3 & {92.3} & 61.8 & {88.9} & 73.2 & 87.8 & 60.0 & {85.5} & {32.3} & 74.8\\  
		SimNet-Res152 \cite{pinheiro2018unsupervised} & {94.3} & 82.3 & 73.5 & 47.2 & 87.9 & 49.2 & 75.1 & 79.7 & 85.3 & 68.5 & 81.1 & 50.3 & 72.9\\
		GTA-Res152 \cite{sankaranarayanan2018generate} & - & - & - & - & - & - & - & - & - & - & - & - & 77.1\\
		\hline
		CBST~\cite{zou2018unsupervised} &87.2 & 78.8 & 56.5 & 55.4 & 85.1 & 79.2 & 83.8 &  77.7 & 82.8 & 88.8 & 69.0 & 72.0 & 76.4\\
        \textbf{+CaCo} &90.7	&80.8	&79.4	&57.0	&89.2	&88.6	&82.4	&79.0	&87.9	&87.9	&87.0	&65.9	&81.3
        \\\hline
		CRST~\cite{zou2019confidence} & 88.0 & 79.2 & 61.0 & 60.0 & 87.5 & 81.4 & 86.3 & 78.8 & 85.6 & 86.6 & 73.9 &   68.8 &78.1\\
        \textbf{+CaCo} &91.4	&80.6	&80.0	&56.5	&89.5	&89.4	&82.8	&79.9	&88.8	&86.8	&87.3	&66.0	&81.6
        \\\hline
		\textbf{CaCo} &90.4	&80.7	&78.8	&57.0	&88.9	&87.0	&81.3	&79.4	&88.7	&88.1	&86.8	&63.9	&80.9
		\\\hline
	\end{tabular}
	}
	\caption{
    Experiments over domain adaptive image classification task VisDA17.
    }
	\label{table:visda17}
\end{table*}

\renewcommand\arraystretch{1.2}
\begin{table}[h]
	\centering
	\resizebox{0.98\linewidth}{!}{
	\begin{tabular}{c|cccccc|c}
		\hline
		Method & A$\rightarrow$W & D$\rightarrow$W & W$\rightarrow$D & A$\rightarrow$D & D$\rightarrow$A & W$\rightarrow$A & Mean\\
		\hline
		Baseline \cite{he2016deep} & 68.4 & 96.7 & 99.3 & 68.9 & 62.5 & 60.7 & 76.1\\\hline
		DAN \cite{long2015learning} & 80.5 & 97.1 & 99.6 & 78.6 & 63.6 & 62.8 & 80.4\\ 
		RTN \cite{long2016unsupervised} & 84.5 & 96.8 & 99.4 & 77.5 & 66.2 & 64.8 & 81.6\\
		DANN \cite{ganin2016domain} & 82.0 & 96.9 & 99.1 & 79.7 & 68.2 & 67.4 & 82.2\\
		ADDA \cite{tzeng2017adversarial} & 86.2 & 96.2 & 98.4 & 77.8 & 69.5 & 68.9 & 82.9\\
		JAN \cite{long2017deep} & 85.4 & 97.4 & 99.8 & 84.7 & 68.6 & 70.0 & 84.3\\
		GTA \cite{sankaranarayanan2018generate} & {89.5} & 97.9 & 99.8 & 87.7 & 72.8 & 71.4 & 86.5\\
		\hline
		CBST~\cite{zou2018unsupervised} & 87.8 & 98.5 & {100.0} & 86.5 & 71.2 & 70.9 & 85.8\\
        \textbf{+CaCo} &90.3	&98.6	&100.0	&92.4	&73.2	&72.8	&87.9
        \\\hline
		CRST~\cite{zou2019confidence} & 89.4 & {98.9} & {100.0 } & 88.7 & 72.6 & 70.9 & 86.8\\
        \textbf{+CaCo} &90.4	&98.9	&100.0	&92.8	&73.7	&72.5	&88.1
        \\\hline
		\textbf{CaCo} &89.7	&98.4	&100.0	&91.7	&73.1	&72.8	&87.6
		\\\hline
	\end{tabular}
	}
\caption{Experiments over domain adaptive image classification task Office-31.
    }
	\label{table:office}
\end{table}

Table~\ref{table:synthia2city} reports semantic segmentation results on the task SYNTHIA $\rightarrow$ Cityscapes. It can be observed that CaCo achieves comparable performance with the highly-optimized state-of-the-art methods, and it boosts their performance (denoted by ``+CaCo'') as well.

\subsection{UDA for Object Detection}
\label{exp_uda_obj}

Tables~\ref{table:det_city2fog} and~\ref{table:det_city2BDD} report object detection experiments on Cityscapes $\rightarrow$ Foggy Cityscapes and Cityscapes $\rightarrow$ BDD100k, respectively. It can be observed that CaCo outperforms the highly-optimized state-of-the-art methods~\cite{saito2019strong,xu2020exploring} clearly. In addition, incorporating CaCo into state-of-the-art methods boosts the detection performance consistently across the two tasks. 

\subsection{UDA for Image Classification}
\label{exp_uda_cla}

Tables~\ref{table:visda17} and ~\ref{table:office} report image classification experiments on VisDA17 and Office-31, respectively. It can be observed that CaCo outperforms state-of-the-art methods clearly. In addition, incorporating CaCo into state-of-the-art methods boosts the image classification consistently in both tasks. 

\subsection{Discussion}
\label{exp_discussion}

\textbf{Generalization across visual recognition tasks:} We study the generalization of the proposed CaCo by evaluating it over three representative visual UDA tasks on \textit{segmentation}, \textit{detection} and \textit{classification}. Experimental results in Tables~\ref{table:gta2city}-~\ref{table:office} show that CaCo achieves competitive performance consistently across all the visual tasks.

\textbf{Complementarity studies:} We study the synergetic benefits of the proposed CaCo by incorporating it into existing UDA methods. Experiments in Tables~\ref{table:gta2city}-~\ref{table:office} (the rows with `+CaCo’) show that CaCo when incorporated improves all existing methods consistently across different visual tasks.

\textbf{Comparisons with existing unsupervised representation learning methods:} We compared CaCo with unsupervised representation learning methods over the UDA task. Most existing methods achieve unsupervised representation learning through certain pretext tasks, such as instance contrastive learning~\cite{chen2020simple,hadsell2006dimensionality,oord2018representation, ye2019unsupervised,bachman2019learning,henaff2020data,wu2018unsupervised,he2020momentum,chen2020improved}, patch ordering~\cite{noroozi2016unsupervised}, rotation prediction~\cite{gidaris2018unsupervised}, and denoising/context/colorization auto-encoders ~\cite{pathak2016context,zhang2016colorful,zhang2017split}. The experiments (shown in Appendix) over the UDA task GTA$\rightarrow$Cityscapes show that existing unsupervised representation learning does not perform well in the UDA task. The major reason is that these methods were designed to learn instance-discriminative representations without considering semantic priors and domain gaps. CaCo also performs unsupervised learning but works for UDA effectively, largely because it learns category-discriminative yet domain-invariant representations which is essential to various visual UDA tasks.

\textbf{Parameter studies:} The parameter $M$ (in the proposed CaCo) controls the length (or size) of the categorical dictionary. We studied $M$ by changing it from $50$ to $150$ with a step of $25$. The experiments (shown in Appendix) over the UDA segmentation task GTA $\rightarrow$ Cityscapes demonstrate that $M$ does not affect UDA clearly while it changes from $50$ to $150$.

\textbf{Generalization across different learning setups:} We studied the scalability of the proposed CaCo from the view of learning setups. Specifically, we evaluated CaCo over a variety of tasks that involve unlabeled data learning and certain semantic priors such as \textit{unsupervised model adaptation}, \textit{partial-set UDA} and \textit{open-set UDA}. Experiments (in Appendix) show that CaCo achieve competitive performance consistently across all the tasks.

\textbf{Category-aware dictionary:} We studied three variant designs of the proposed category-aware dictionary: 1) Assign all keys with the same temperature; 2) Using two individual dictionaries (for source and target data) instead of a single domain-mixed dictionary; 3) Update the dictionary by memory bank~\cite{wu2018unsupervised} or current mini-batch~\cite{chen2020simple}. Experiments (in Appendix) verify the superiority of the design as described in this paper.

\section{Conclusion}

This paper presents CaCo, a category contrast technique that introduces a generic category contrastive loss that can work for various visual UDA tasks effectively. We construct a semantics-aware dictionary with samples from both source and target domains where each target sample is assigned a (pseudo) category label based on the category priors of source samples. This allows category contrastive learning (between target queries and the category-level dictionary) for category-discriminative yet domain-invariant feature representations: samples of the same category (from either source or target domain) are pulled close together while those of different categories are pushed away simultaneously. Extensive experiments over multiple visual tasks ($e.g.$, segmentation, classification and detection) show that the simple implementation of CaCo achieves superior performance as compared with highly-optimized state-of-the-art methods. In addition, we demonstrate that CaCo is also complementary to existing UDA methods and generalizable to other learning setups such as unsupervised model adaptation, open-/partial-set adaptation etc.

\section*{Acknowledgement}
This research was conducted at Singtel Cognitive and Artificial Intelligence Lab for Enterprises (SCALE@NTU), which is a collaboration between Singapore Telecommunications Limited (Singtel) and Nanyang Technological University (NTU) that is supported by A*STAR under its Industry Alignment Fund (LOA Award number: I1701E0013).

{\small
\bibliographystyle{ieee_fullname}
\bibliography{egbib}

\begin{thebibliography}{10}\itemsep=-1pt

\bibitem{alonso2021semi}
Inigo Alonso, Alberto Sabater, David Ferstl, Luis Montesano, and Ana~C Murillo.
\newblock Semi-supervised semantic segmentation with pixel-level contrastive
  learning from a class-wise memory bank.
\newblock In {\em Proceedings of the IEEE/CVF International Conference on
  Computer Vision}, pages 8219--8228, 2021.

\bibitem{bachman2019learning}
Philip Bachman, R~Devon Hjelm, and William Buchwalter.
\newblock Learning representations by maximizing mutual information across
  views.
\newblock {\em arXiv preprint arXiv:1906.00910}, 2019.

\bibitem{bottou2010large}
L{\'e}on Bottou.
\newblock Large-scale machine learning with stochastic gradient descent.
\newblock In {\em Proceedings of COMPSTAT'2010}, pages 177--186. Springer,
  2010.

\bibitem{chen2017deeplab}
Liang-Chieh Chen, George Papandreou, Iasonas Kokkinos, Kevin Murphy, and Alan~L
  Yuille.
\newblock Deeplab: Semantic image segmentation with deep convolutional nets,
  atrous convolution, and fully connected crfs.
\newblock {\em IEEE transactions on pattern analysis and machine intelligence},
  40(4):834--848, 2017.

\bibitem{chen2020simple}
Ting Chen, Simon Kornblith, Mohammad Norouzi, and Geoffrey Hinton.
\newblock A simple framework for contrastive learning of visual
  representations.
\newblock In {\em International conference on machine learning}, pages
  1597--1607. PMLR, 2020.

\bibitem{chen2020improved}
Xinlei Chen, Haoqi Fan, Ross Girshick, and Kaiming He.
\newblock Improved baselines with momentum contrastive learning.
\newblock {\em arXiv preprint arXiv:2003.04297}, 2020.

\bibitem{chen2018domain}
Yuhua Chen, Wen Li, Christos Sakaridis, Dengxin Dai, and Luc Van~Gool.
\newblock Domain adaptive faster r-cnn for object detection in the wild.
\newblock In {\em Proceedings of the IEEE conference on computer vision and
  pattern recognition}, pages 3339--3348, 2018.

\bibitem{chen2019crdoco}
Yun-Chun Chen, Yen-Yu Lin, Ming-Hsuan Yang, and Jia-Bin Huang.
\newblock Crdoco: Pixel-level domain transfer with cross-domain consistency.
\newblock In {\em Proceedings of the IEEE Conference on Computer Vision and
  Pattern Recognition}, pages 1791--1800, 2019.

\bibitem{cordts2016cityscapes}
Marius Cordts, Mohamed Omran, Sebastian Ramos, Timo Rehfeld, Markus Enzweiler,
  Rodrigo Benenson, Uwe Franke, Stefan Roth, and Bernt Schiele.
\newblock The cityscapes dataset for semantic urban scene understanding.
\newblock In {\em Proceedings of the IEEE conference on computer vision and
  pattern recognition}, pages 3213--3223, 2016.

\bibitem{cui2021genco}
Kaiwen Cui, Jiaxing Huang, Zhipeng Luo, Gongjie Zhang, Fangneng Zhan, and
  Shijian Lu.
\newblock Genco: Generative co-training for generative adversarial networks
  with limited data.
\newblock {\em arXiv preprint arXiv:2110.01254}, 2021.

\bibitem{ganin2016domain}
Yaroslav Ganin, Evgeniya Ustinova, Hana Ajakan, Pascal Germain, Hugo
  Larochelle, Fran{\c{c}}ois Laviolette, Mario Marchand, and Victor Lempitsky.
\newblock Domain-adversarial training of neural networks.
\newblock {\em The Journal of Machine Learning Research}, 17(1):2096--2030,
  2016.

\bibitem{gidaris2018unsupervised}
Spyros Gidaris, Praveer Singh, and Nikos Komodakis.
\newblock Unsupervised representation learning by predicting image rotations.
\newblock {\em arXiv preprint arXiv:1803.07728}, 2018.

\bibitem{grandvalet2005semi}
Yves Grandvalet and Yoshua Bengio.
\newblock Semi-supervised learning by entropy minimization.
\newblock In {\em Advances in neural information processing systems}, pages
  529--536, 2005.

\bibitem{guan2021scale}
Dayan Guan, Jiaxing Huang, Shijian Lu, and Aoran Xiao.
\newblock Scale variance minimization for unsupervised domain adaptation in
  image segmentation.
\newblock {\em Pattern Recognition}, 112:107764, 2021.

\bibitem{guan2021domain}
Dayan Guan, Jiaxing Huang, Aoran Xiao, and Shijian Lu.
\newblock Domain adaptive video segmentation via temporal consistency
  regularization.
\newblock In {\em Proceedings of the IEEE/CVF International Conference on
  Computer Vision}, pages 8053--8064, 2021.

\bibitem{guan2021uncertainty}
Dayan Guan, Jiaxing Huang, Aoran Xiao, Shijian Lu, and Yanpeng Cao.
\newblock Uncertainty-aware unsupervised domain adaptation in object detection.
\newblock {\em IEEE Transactions on Multimedia}, 2021.

\bibitem{gunel2020supervised}
Beliz Gunel, Jingfei Du, Alexis Conneau, and Ves Stoyanov.
\newblock Supervised contrastive learning for pre-trained language model
  fine-tuning.
\newblock {\em arXiv preprint arXiv:2011.01403}, 2020.

\bibitem{hadsell2006dimensionality}
Raia Hadsell, Sumit Chopra, and Yann LeCun.
\newblock Dimensionality reduction by learning an invariant mapping.
\newblock In {\em 2006 IEEE Computer Society Conference on Computer Vision and
  Pattern Recognition (CVPR'06)}, volume~2, pages 1735--1742. IEEE, 2006.

\bibitem{he2020momentum}
Kaiming He, Haoqi Fan, Yuxin Wu, Saining Xie, and Ross Girshick.
\newblock Momentum contrast for unsupervised visual representation learning.
\newblock In {\em Proceedings of the IEEE/CVF Conference on Computer Vision and
  Pattern Recognition}, pages 9729--9738, 2020.

\bibitem{he2016deep}
Kaiming He, Xiangyu Zhang, Shaoqing Ren, and Jian Sun.
\newblock Deep residual learning for image recognition.
\newblock In {\em Proceedings of the IEEE conference on computer vision and
  pattern recognition}, pages 770--778, 2016.

\bibitem{he2019multi}
Zhenwei He and Lei Zhang.
\newblock Multi-adversarial faster-rcnn for unrestricted object detection.
\newblock In {\em Proceedings of the IEEE/CVF International Conference on
  Computer Vision}, pages 6668--6677, 2019.

\bibitem{henaff2020data}
Olivier Henaff.
\newblock Data-efficient image recognition with contrastive predictive coding.
\newblock In {\em International Conference on Machine Learning}, pages
  4182--4192. PMLR, 2020.

\bibitem{hsu2020every}
Cheng-Chun Hsu, Yi-Hsuan Tsai, Yen-Yu Lin, and Ming-Hsuan Yang.
\newblock Every pixel matters: Center-aware feature alignment for domain
  adaptive object detector.
\newblock In {\em European Conference on Computer Vision}, pages 733--748.
  Springer, 2020.

\bibitem{huang2021cross}
Jiaxing Huang, Dayan Guan, Aoran Xiao, and Shijian Lu.
\newblock Cross-view regularization for domain adaptive panoptic segmentation.
\newblock In {\em Proceedings of the IEEE/CVF Conference on Computer Vision and
  Pattern Recognition}, pages 10133--10144, 2021.

\bibitem{huang2021fsdr}
Jiaxing Huang, Dayan Guan, Aoran Xiao, and Shijian Lu.
\newblock Fsdr: Frequency space domain randomization for domain generalization.
\newblock In {\em Proceedings of the IEEE/CVF Conference on Computer Vision and
  Pattern Recognition}, pages 6891--6902, 2021.

\bibitem{huang2021model}
Jiaxing Huang, Dayan Guan, Aoran Xiao, and Shijian Lu.
\newblock Model adaptation: Historical contrastive learning for unsupervised
  domain adaptation without source data.
\newblock {\em Advances in Neural Information Processing Systems}, 34, 2021.

\bibitem{huang2021rda}
Jiaxing Huang, Dayan Guan, Aoran Xiao, and Shijian Lu.
\newblock Rda: Robust domain adaptation via fourier adversarial attacking.
\newblock In {\em Proceedings of the IEEE/CVF International Conference on
  Computer Vision}, pages 8988--8999, 2021.

\bibitem{huang2022multi}
Jiaxing Huang, Dayan Guan, Aoran Xiao, and Shijian Lu.
\newblock Multi-level adversarial network for domain adaptive semantic
  segmentation.
\newblock {\em Pattern Recognition}, 123:108384, 2022.

\bibitem{huang2020contextual}
Jiaxing Huang, Shijian Lu, Dayan Guan, and Xiaobing Zhang.
\newblock Contextual-relation consistent domain adaptation for semantic
  segmentation.
\newblock In {\em European Conference on Computer Vision}, pages 705--722.
  Springer, 2020.

\bibitem{kang2019contrastive}
Guoliang Kang, Lu Jiang, Yi Yang, and Alexander~G Hauptmann.
\newblock Contrastive adaptation network for unsupervised domain adaptation.
\newblock In {\em Proceedings of the IEEE Conference on Computer Vision and
  Pattern Recognition}, pages 4893--4902, 2019.

\bibitem{khosla2020supervised}
Prannay Khosla, Piotr Teterwak, Chen Wang, Aaron Sarna, Yonglong Tian, Phillip
  Isola, Aaron Maschinot, Ce Liu, and Dilip Krishnan.
\newblock Supervised contrastive learning.
\newblock {\em arXiv preprint arXiv:2004.11362}, 2020.

\bibitem{kim2020learning}
Myeongjin Kim and Hyeran Byun.
\newblock Learning texture invariant representation for domain adaptation of
  semantic segmentation.
\newblock {\em arXiv preprint arXiv:2003.00867}, 2020.

\bibitem{kim2019diversify}
Taekyung Kim, Minki Jeong, Seunghyeon Kim, Seokeon Choi, and Changick Kim.
\newblock Diversify and match: A domain adaptive representation learning
  paradigm for object detection.
\newblock In {\em Proceedings of the IEEE/CVF Conference on Computer Vision and
  Pattern Recognition}, pages 12456--12465, 2019.

\bibitem{li2021semantic}
Shuang Li, Binhui Xie, Bin Zang, Chi~Harold Liu, Xinjing Cheng, Ruigang Yang,
  and Guoren Wang.
\newblock Semantic distribution-aware contrastive adaptation for semantic
  segmentation.
\newblock {\em arXiv preprint arXiv:2105.05013}, 2021.

\bibitem{li2019bidirectional}
Yunsheng Li, Lu Yuan, and Nuno Vasconcelos.
\newblock Bidirectional learning for domain adaptation of semantic
  segmentation.
\newblock In {\em Proceedings of the IEEE Conference on Computer Vision and
  Pattern Recognition}, pages 6936--6945, 2019.

\bibitem{long2015learning}
Mingsheng Long, Yue Cao, Jianmin Wang, and Michael Jordan.
\newblock Learning transferable features with deep adaptation networks.
\newblock In {\em International Conference on Machine Learning}, pages 97--105,
  2015.

\bibitem{long2016unsupervised}
Mingsheng Long, Han Zhu, Jianmin Wang, and Michael~I Jordan.
\newblock Unsupervised domain adaptation with residual transfer networks.
\newblock In {\em Advances in Neural Information Processing Systems}, pages
  136--144, 2016.

\bibitem{long2017deep}
Mingsheng Long, Han Zhu, Jianmin Wang, and Michael~I Jordan.
\newblock Deep transfer learning with joint adaptation networks.
\newblock In {\em International conference on machine learning}, pages
  2208--2217. PMLR, 2017.

\bibitem{luo2019taking}
Yawei Luo, Liang Zheng, Tao Guan, Junqing Yu, and Yi Yang.
\newblock Taking a closer look at domain shift: Category-level adversaries for
  semantics consistent domain adaptation.
\newblock In {\em Proceedings of the IEEE Conference on Computer Vision and
  Pattern Recognition}, pages 2507--2516, 2019.

\bibitem{noroozi2016unsupervised}
Mehdi Noroozi and Paolo Favaro.
\newblock Unsupervised learning of visual representations by solving jigsaw
  puzzles.
\newblock In {\em European Conference on Computer Vision}, pages 69--84.
  Springer, 2016.

\bibitem{oord2018representation}
Aaron van~den Oord, Yazhe Li, and Oriol Vinyals.
\newblock Representation learning with contrastive predictive coding.
\newblock {\em arXiv preprint arXiv:1807.03748}, 2018.

\bibitem{pan2020unsupervised}
Fei Pan, Inkyu Shin, Francois Rameau, Seokju Lee, and In~So Kweon.
\newblock Unsupervised intra-domain adaptation for semantic segmentation
  through self-supervision.
\newblock {\em arXiv preprint arXiv:2004.07703}, 2020.

\bibitem{pathak2016context}
Deepak Pathak, Philipp Krahenbuhl, Jeff Donahue, Trevor Darrell, and Alexei~A
  Efros.
\newblock Context encoders: Feature learning by inpainting.
\newblock In {\em Proceedings of the IEEE conference on computer vision and
  pattern recognition}, pages 2536--2544, 2016.

\bibitem{peng2018visda}
Xingchao Peng, Ben Usman, Neela Kaushik, Dequan Wang, Judy Hoffman, and Kate
  Saenko.
\newblock Visda: A synthetic-to-real benchmark for visual domain adaptation.
\newblock In {\em Proceedings of the IEEE Conference on Computer Vision and
  Pattern Recognition Workshops}, pages 2021--2026, 2018.

\bibitem{pinheiro2018unsupervised}
Pedro~O Pinheiro.
\newblock Unsupervised domain adaptation with similarity learning.
\newblock In {\em Proceedings of the IEEE Conference on Computer Vision and
  Pattern Recognition}, pages 8004--8013, 2018.

\bibitem{ren2015faster}
Shaoqing Ren, Kaiming He, Ross Girshick, and Jian Sun.
\newblock Faster r-cnn: Towards real-time object detection with region proposal
  networks.
\newblock In {\em Advances in neural information processing systems}, pages
  91--99, 2015.

\bibitem{richter2016playing}
Stephan~R Richter, Vibhav Vineet, Stefan Roth, and Vladlen Koltun.
\newblock Playing for data: Ground truth from computer games.
\newblock In {\em European conference on computer vision}, pages 102--118.
  Springer, 2016.

\bibitem{ros2016synthia}
German Ros, Laura Sellart, Joanna Materzynska, David Vazquez, and Antonio~M
  Lopez.
\newblock The synthia dataset: A large collection of synthetic images for
  semantic segmentation of urban scenes.
\newblock In {\em Proceedings of the IEEE conference on computer vision and
  pattern recognition}, pages 3234--3243, 2016.

\bibitem{saenko2010adapting}
Kate Saenko, Brian Kulis, Mario Fritz, and Trevor Darrell.
\newblock Adapting visual category models to new domains.
\newblock In {\em European conference on computer vision}, pages 213--226.
  Springer, 2010.

\bibitem{saito2018adversarial}
Kuniaki Saito, Yoshitaka Ushiku, Tatsuya Harada, and Kate Saenko.
\newblock Adversarial dropout regularization.
\newblock {\em International Conference on Learning Representations}, 2017.

\bibitem{saito2019strong}
Kuniaki Saito, Yoshitaka Ushiku, Tatsuya Harada, and Kate Saenko.
\newblock Strong-weak distribution alignment for adaptive object detection.
\newblock In {\em Proceedings of the IEEE/CVF Conference on Computer Vision and
  Pattern Recognition}, pages 6956--6965, 2019.

\bibitem{saito2018maximum}
Kuniaki Saito, Kohei Watanabe, Yoshitaka Ushiku, and Tatsuya Harada.
\newblock Maximum classifier discrepancy for unsupervised domain adaptation.
\newblock In {\em Proceedings of the IEEE Conference on Computer Vision and
  Pattern Recognition}, pages 3723--3732, 2018.

\bibitem{sakaridis2018semantic}
Christos Sakaridis, Dengxin Dai, and Luc Van~Gool.
\newblock Semantic foggy scene understanding with synthetic data.
\newblock {\em International Journal of Computer Vision}, 126(9):973--992,
  2018.

\bibitem{sankaranarayanan2018generate}
Swami Sankaranarayanan, Yogesh Balaji, Carlos~D Castillo, and Rama Chellappa.
\newblock Generate to adapt: Aligning domains using generative adversarial
  networks.
\newblock In {\em Proceedings of the IEEE Conference on Computer Vision and
  Pattern Recognition}, pages 8503--8512, 2018.

\bibitem{saunshi2019theoretical}
Nikunj Saunshi, Orestis Plevrakis, Sanjeev Arora, Mikhail Khodak, and
  Hrishikesh Khandeparkar.
\newblock A theoretical analysis of contrastive unsupervised representation
  learning.
\newblock In {\em International Conference on Machine Learning}, pages
  5628--5637. PMLR, 2019.

\bibitem{simonyan2014very}
Karen Simonyan and Andrew Zisserman.
\newblock Very deep convolutional networks for large-scale image recognition.
\newblock {\em arXiv preprint arXiv:1409.1556}, 2014.

\bibitem{tian2019contrastive}
Yonglong Tian, Dilip Krishnan, and Phillip Isola.
\newblock Contrastive multiview coding.
\newblock {\em arXiv preprint arXiv:1906.05849}, 2019.

\bibitem{tsai2018learning}
Yi-Hsuan Tsai, Wei-Chih Hung, Samuel Schulter, Kihyuk Sohn, Ming-Hsuan Yang,
  and Manmohan Chandraker.
\newblock Learning to adapt structured output space for semantic segmentation.
\newblock In {\em Proceedings of the IEEE Conference on Computer Vision and
  Pattern Recognition}, pages 7472--7481, 2018.

\bibitem{tsai2019domain}
Yi-Hsuan Tsai, Kihyuk Sohn, Samuel Schulter, and Manmohan Chandraker.
\newblock Domain adaptation for structured output via discriminative patch
  representations.
\newblock In {\em Proceedings of the IEEE International Conference on Computer
  Vision}, pages 1456--1465, 2019.

\bibitem{tschannen2019mutual}
Michael Tschannen, Josip Djolonga, Paul~K Rubenstein, Sylvain Gelly, and Mario
  Lucic.
\newblock On mutual information maximization for representation learning.
\newblock {\em arXiv preprint arXiv:1907.13625}, 2019.

\bibitem{tzeng2017adversarial}
Eric Tzeng, Judy Hoffman, Kate Saenko, and Trevor Darrell.
\newblock Adversarial discriminative domain adaptation.
\newblock In {\em Proceedings of the IEEE Conference on Computer Vision and
  Pattern Recognition}, pages 7167--7176, 2017.

\bibitem{vu2019advent}
Tuan-Hung Vu, Himalaya Jain, Maxime Bucher, Matthieu Cord, and Patrick
  P{\'e}rez.
\newblock Advent: Adversarial entropy minimization for domain adaptation in
  semantic segmentation.
\newblock In {\em Proceedings of the IEEE Conference on Computer Vision and
  Pattern Recognition}, pages 2517--2526, 2019.

\bibitem{wang2021domain}
Qin Wang, Dengxin Dai, Lukas Hoyer, Luc Van~Gool, and Olga Fink.
\newblock Domain adaptive semantic segmentation with self-supervised depth
  estimation.
\newblock In {\em Proceedings of the IEEE/CVF International Conference on
  Computer Vision}, pages 8515--8525, 2021.

\bibitem{wang2021exploring}
Wenguan Wang, Tianfei Zhou, Fisher Yu, Jifeng Dai, Ender Konukoglu, and Luc
  Van~Gool.
\newblock Exploring cross-image pixel contrast for semantic segmentation.
\newblock In {\em Proceedings of the IEEE/CVF International Conference on
  Computer Vision}, pages 7303--7313, 2021.

\bibitem{wang2015unsupervised}
Xiaolong Wang and Abhinav Gupta.
\newblock Unsupervised learning of visual representations using videos.
\newblock In {\em Proceedings of the IEEE international conference on computer
  vision}, pages 2794--2802, 2015.

\bibitem{wang2020differential}
Zhonghao Wang, Mo Yu, Yunchao Wei, Rogerior Feris, Jinjun Xiong, Wen-mei Hwu,
  Thomas~S Huang, and Honghui Shi.
\newblock Differential treatment for stuff and things: A simple unsupervised
  domain adaptation method for semantic segmentation.
\newblock {\em arXiv preprint arXiv:2003.08040}, 2020.

\bibitem{wu2018unsupervised}
Zhirong Wu, Yuanjun Xiong, Stella~X Yu, and Dahua Lin.
\newblock Unsupervised feature learning via non-parametric instance
  discrimination.
\newblock In {\em Proceedings of the IEEE Conference on Computer Vision and
  Pattern Recognition}, pages 3733--3742, 2018.

\bibitem{xiao2022unsupervised}
Aoran Xiao, Jiaxing Huang, Dayan Guan, and Shijian Lu.
\newblock Unsupervised representation learning for point clouds: A survey.
\newblock {\em arXiv preprint arXiv:2202.13589}, 2022.

\bibitem{Xie_2019_ICCV}
Rongchang Xie, Fei Yu, Jiachao Wang, Yizhou Wang, and Li Zhang.
\newblock Multi-level domain adaptive learning for cross-domain detection.
\newblock In {\em Proceedings of the IEEE/CVF International Conference on
  Computer Vision (ICCV) Workshops}, Oct 2019.

\bibitem{xu2020exploring}
Chang-Dong Xu, Xing-Ran Zhao, Xin Jin, and Xiu-Shen Wei.
\newblock Exploring categorical regularization for domain adaptive object
  detection.
\newblock In {\em Proceedings of the IEEE/CVF Conference on Computer Vision and
  Pattern Recognition}, pages 11724--11733, 2020.

\bibitem{yang2020fda}
Yanchao Yang and Stefano Soatto.
\newblock Fda: Fourier domain adaptation for semantic segmentation.
\newblock In {\em Proceedings of the IEEE/CVF Conference on Computer Vision and
  Pattern Recognition}, pages 4085--4095, 2020.

\bibitem{ye2019unsupervised}
Mang Ye, Xu Zhang, Pong~C Yuen, and Shih-Fu Chang.
\newblock Unsupervised embedding learning via invariant and spreading instance
  feature.
\newblock In {\em Proceedings of the IEEE/CVF Conference on Computer Vision and
  Pattern Recognition}, pages 6210--6219, 2019.

\bibitem{yu2018bdd100k}
Fisher Yu, Wenqi Xian, Yingying Chen, Fangchen Liu, Mike Liao, Vashisht
  Madhavan, and Trevor Darrell.
\newblock Bdd100k: A diverse driving video database with scalable annotation
  tooling.
\newblock {\em arXiv preprint arXiv:1805.04687}, 2(5):6, 2018.

\bibitem{zhang2021spectral}
Jingyi Zhang, Jiaxing Huang, and Shijian Lu.
\newblock Spectral unsupervised domain adaptation for visual recognition.
\newblock In {\em Proceedings of the IEEE/CVF Conference on Computer Vision and
  Pattern Recognition}, pages 0--0, 2022.

\bibitem{zhang2021detr}
Jingyi Zhang, Jiaxing Huang, Zhipeng Luo, Gongjie Zhang, and Shijian Lu.
\newblock Da-detr: Domain adaptive detection transformer by hybrid attention.
\newblock {\em arXiv preprint arXiv:2103.17084}, 2021.

\bibitem{zhang2021proda}
Pan Zhang, Bo Zhang, Ting Zhang, Dong Chen, Yong Wang, and Fang Wen.
\newblock Prototypical pseudo label denoising and target structure learning for
  domain adaptive semantic segmentation.
\newblock In {\em Proceedings of the IEEE/CVF Conference on Computer Vision and
  Pattern Recognition}, pages 12414--12424, 2021.

\bibitem{zhang2016colorful}
Richard Zhang, Phillip Isola, and Alexei~A Efros.
\newblock Colorful image colorization.
\newblock In {\em European conference on computer vision}, pages 649--666.
  Springer, 2016.

\bibitem{zhang2017split}
Richard Zhang, Phillip Isola, and Alexei~A Efros.
\newblock Split-brain autoencoders: Unsupervised learning by cross-channel
  prediction.
\newblock In {\em Proceedings of the IEEE Conference on Computer Vision and
  Pattern Recognition}, pages 1058--1067, 2017.

\bibitem{zhu2019adapting}
Xinge Zhu, Jiangmiao Pang, Ceyuan Yang, Jianping Shi, and Dahua Lin.
\newblock Adapting object detectors via selective cross-domain alignment.
\newblock In {\em Proceedings of the IEEE/CVF Conference on Computer Vision and
  Pattern Recognition}, pages 687--696, 2019.

\bibitem{zou2019confidence}
Yang Zou, Zhiding Yu, Xiaofeng Liu, BVK Kumar, and Jinsong Wang.
\newblock Confidence regularized self-training.
\newblock In {\em Proceedings of the IEEE International Conference on Computer
  Vision}, pages 5982--5991, 2019.

\bibitem{zou2018unsupervised}
Yang Zou, Zhiding Yu, BVK Vijaya~Kumar, and Jinsong Wang.
\newblock Unsupervised domain adaptation for semantic segmentation via
  class-balanced self-training.
\newblock In {\em Proceedings of the European Conference on Computer Vision
  (ECCV)}, pages 289--305, 2018.

\end{thebibliography}
}

\end{document}